\documentclass[journal]{IEEEtran}

\ifCLASSOPTIONcompsoc
  \usepackage[nocompress]{cite}
\else
  \usepackage{cite}
\fi

\usepackage{times}
\usepackage{amsmath}
\usepackage{epsfig}
\usepackage{graphicx}
\usepackage{amsmath}
\usepackage{amssymb}
\usepackage{subfigure}
\usepackage{multirow}
\usepackage{algorithmicx}
\usepackage{color, colortbl}
\usepackage[lined, ruled]{algorithm2e}
\usepackage[breaklinks=true,bookmarks=false]{hyperref}

\makeatletter
\DeclareRobustCommand\onedot{\futurelet\@let@token\@onedot}
\def\@onedot{\ifx\@let@token.\else.\null\fi\xspace}

\def\etal{\emph{et al}\onedot}

\newcommand{\tb}[1]{\textcolor{blue}{#1}}

\newcommand{\tR}[1]{{\textbf{#1}}} 
\newcommand{\tbf}[1]{\textbf{#1}}
\newcommand{\ti}[1]{\texttt{#1}}

\makeatother

\begin{document}
\title{PaMM: Pose-aware Multi-shot Matching for Improving Person Re-identification}

\author{Yeong-Jun~Cho and~Kuk-Jin~Yoon \\ Computer Vision Laboratory, GIST, South Korea
\IEEEcompsocitemizethanks{\IEEEcompsocthanksitem Yeong-Jun Cho and Kuk-Jin Yoon are with the School of Information and Communications, Gwangju Institute of Science and Technology, South Korea.\protect\\
E-mail: kjyoon@gist.ac.kr}
}




\IEEEtitleabstractindextext{%
\begin{abstract}
Person re-identification is the problem of recognizing people across different images or videos with non-overlapping views.
Although there has been much progress in person re-identification over the last decade, it remains a challenging task because appearances of people can seem extremely different across diverse camera viewpoints and person poses. 
In this paper, we propose a novel framework for person re-identification by analyzing camera viewpoints and person poses in a so-called Pose-aware Multi-shot Matching~(PaMM), which robustly estimates people's poses and efficiently conducts multi-shot matching based on pose information. Experimental results using public person re-identification datasets show that the proposed methods outperform state-of-the-art methods and are promising for person re-identification from diverse viewpoints and pose variances.
\end{abstract}

}

\maketitle

\IEEEdisplaynontitleabstractindextext
\IEEEpeerreviewmaketitle

\ifCLASSOPTIONcompsoc

\fi

\section{Introduction}\label{sec:introduction}

\IEEEPARstart{A} huge number of surveillance cameras have been installed in public places (\textit{e.g.} offices, stations, airports, and streets) in recent year to closely monitor scenes and give early warnings of events such as accidents and crimes. 
However, dealing with large camera networks requires a lot of human effort.
Automatic person re-identification tasks that can associate people across images from non-overlapping cameras have been widely utilized to reduce the required human effort.

\begin{figure}[t]
	\includegraphics[width=1\columnwidth]{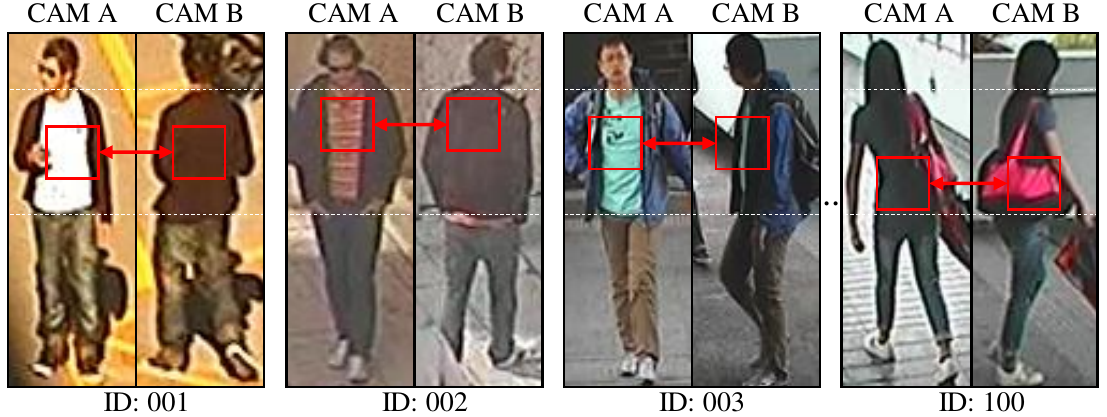}  
	\caption{Challenging in person re-identification due to person appearance changes. Person appearance changes depending on the camera viewpoint and the pose of a person. Pairs of red squares have same locations but show different appearances due to pose variations of people.}
	\label{fig:chal_reid} 
\end{figure}

Most previous works have generally relied on people's appearances such as color, shape, and texture to re-identify them, since there is no continuity between non-overlapping cameras in terms of time and space.
Thus, many works have focused on appearance modeling and learning such as via feature descriptor extraction~\cite{farenzena2010person,zhao2014learning}, metric learning~\cite{koestinger2012large, roth2014mahalanobis}, and saliency learning~\cite{zhao2013unsupervised} methods for efficient re-identification.	
Unfortunately, a person's appearance can change considerably across images depending on a camera's viewpoint and a person's pose as shown in Fig.~\ref{fig:chal_reid}; thus, person re-identification tasks that only rely on appearance are very challenging.
Nonetheless, many previous re-identification frameworks~\cite{koestinger2012large, roth2014mahalanobis, zhao2013unsupervised} have commonly adopted single-shot matching methods that utilize a single appearance for each person to measure the similarity (\textit{or} difference) between a pair of person image patches. However, it is difficult to identify people with single-shot appearance matching because of the people's aforementioned severe appearance changes.
Several multi-shot matching methods~\cite{farenzena2010person, wang2014person, limulti} have been proposed in recent years to overcome the limitation of single-shot matching; however, the ambiguities that arise due to the viewpoint and pose variations remain.

In real world surveillance scenarios, each person provides multiple observations along a moving path. Therefore, we can exploit plenty of appearances for  re-identification tasks.
Furthermore, surveillance videos contain scene structures and scene contexts such as a ground plane of a scene, a person's trajectory, etc. 
In practice, it is possible to estimate the camera viewpoint from the scene information via human height-based auto-calibrations~\cite{kusakunniran2009direct,lv2006camera} and vanishing point-based auto-calibrations~\cite{orghidan2012camera}; then the difficulties in person re-identification become more tractable.

This paper proposes a novel framework for person re-identification by analyzing camera viewpoints and person poses called Pose-aware Multi-shot Matching~(PaMM). First camera viewpoints are calibrated and poeple's poses are robustly estimated based on the proposed pose estimation method. 
We then generate a multi-pose model that contains four feature descriptor groups extracted from four image clusters grouped by person poses (\textit{i.e.} \textit{f}ront, \textit{r}ight, \textit{b}ack, and \textit{l}eft). 
After generating multi-pose models, we calculate matching scores between multi-pose models in a weighted summation manner based on pre-trained matching weights.
The proposed person re-identification framework permits the exploitation of additional cues such as person poses and 3D scene information, which makes the person re-identification problem more tractable.

To validate our methods, we extensively evaluate the performance of the proposed methods and other state-of-the-art methods that use public person re-identification datasets \texttt{3DPeS}~\cite{baltieri2011_308}, \texttt{PRID 2011}~\cite{hirzer11a} and \texttt{iLIDS-Vid}~\cite{wang2014person}. Experimental results show that the proposed framework is promising for person re-identification from diverse viewpoint and pose variations and outperforms other state-of-the-art methods.
For reproducibility, the PaMM code is openly available to the public at: \url{https://cvl.gist.ac.kr/pamm/}.

The main ideas of this work are simple but very effective. In addition, our method can flexibly adopt any existing person re-identification methods such as feature descriptor extraction~\cite{farenzena2010person,zhao2014learning} and metric learning~\cite{davis2007information, koestinger2012large, weinberger2005distance} methods as the baseline in our re-identification framework.
This is the first attempt to exploit viewpoint and pose information for \emph{multi-shot} person re-identification to the best of our knowledge.

The rest of the paper is organized as follows: Section~\ref{sec:preivous} summarizes previous person re-identification works. 
Section~\ref{sec:motiv}, explains the motivation behind this work. We then describe our proposed methods in Section~\ref{sec:proposed}. 
The datasets and evaluation methodology used are described in Section~\ref{sec:data_metho}. 
The experimental results are reported in Section~\ref{sec:exp} and we conclude this paper in Section~\ref{sec:conclusion}.

	\section{Previous Works}
	\label{sec:preivous}
	
	We have classified previous person re-identification methods into single-shot matching methods and multi-shot matching methods and briefly reviewed them. Single-shot matching methods only use a single appearance of each person to find people correspondences between two different cameras, whereas the multi-shot matching methods exploit multiple appearances to find the correspondences.
	
	\subsection{Single-shot matching methods}
	
	\label{subsec:single_shot}
	
	Most works that attempt to re-identify people across non-overlapping cameras generally rely on poeple's appearance since there is no spatiotemporal continuity; we cannot fully utilize the motion or spatial information of a person for their re-identification.
	Therefore, most works have focused on appearance-based techniques such as feature descriptor extraction and metric learning methods for efficient person re-identification.
	
	Regarding feature extraction methods, Farenzena~\etal~\cite{farenzena2010person} proposed the symmetry-driven accumulation of local features that are extracted based on the principles of the symmetry and asymmetry of the human body. 
	This method exploits the human body model which is robust to human pose variations.
	Feature extraction methods that select or weight discriminative features have been proposed in~\cite{liu2012person,zhao2014learning}. These methods enable us to adaptively exploit features depending on the person's appearance. 
	Regarding metric learning, several methods have been proposed such as KISSME~\cite{koestinger2012large}, LMNN-R~\cite{dikmen2011pedestrian}, and applied to the re-identification problem.
	Some works \cite{koestinger2012large,roth2014mahalanobis} have extensively evaluated and compared several metric learning methods (\textit{e.g.} ITML~\cite{davis2007information}, KISSME~\cite{koestinger2012large}, LMNN~\cite{weinberger2005distance} and Mahalnobis~\cite{roth2014mahalanobis}) and shown the effectiveness of metric learning for re-identification.
	Similar to metric learning methods, a saliency learning method was also proposed by R. Zhao~\etal~\cite{zhao2013unsupervised} that learned saliency for handling severe appearance changes.
	Recently, many person re-identification methods have been proposed that are based on learning deep convolutional neural network~(CNN)~\cite{su2016deep} and Siamese convolutional network~\cite{ahmed2015improved, yi2014deep,wang2016joint} for simultaneously learning both features and metrics. In addition, \cite{liao2015person} proposed both feature descriptor extraction and metric learning methods for re-identification.
	
	However, a few works~\cite{bak2015person,wu2015viewpoint} that use person pose for re-identification have been proposed very recently.
	Bak \etal~\cite{bak2015person} proposed learning a generic metric pool that consists of metrics, each of which are learned to match specific pairs of poses. 
	Wu. \etal~\cite{wu2015viewpoint} proposed person re-identification involving human appearance modeling using pose priors and person-specific feature learning. Although these methods utilized pose priors for person re-identification, they consider single-shot matching that recognizes people using a single appearance, which has difficulties for handling diverse appearance changes. This paper proposes a person re-identification framework that uses pose cues for efficient \emph{multi-shot matching}.

	\subsection{Multi-shot matching methods}  
	
	Several multi-shot matching methods that have sought to overcome the limitations of single-shot matching methods have been proposed in recent years. 
	Li~\etal~\cite{li2015multi} proposed a random forest-based person re-identification that exploits multiple appearances. They calculated similarity scores between two multi-shot sets and averaged them into a final similarity score.
	Farenzena~\etal~\cite{farenzena2010person} also provided multi-shot matching results by comparing each possible pair of histograms between different signatures (a set of appearances) and selecting the lowest obtained distance for the final matching score. 
	Similarly, Su~\etal\cite{su2016deep} and Zheng~\etal\cite{zheng2015scalable} exploited multiple queries for re-identification. Instead of the multi-shot matching strategies such as \cite{farenzena2010person,li2015multi}, they merged multiple queries (\textit{i.e.} multi-shot) into a single query and performed re-identification using the merged queries.

	Wang~\etal~\cite{wang2014person,wang2016person} proposed video ranking methods for multi-shot matching that automatically selected discriminative video fragments and learned a video ranking function. You~\etal~\cite{you2016top} proposed a top-push distance learning model for efficiently matching video features of people.
	Similarly, Liu~\etal~\cite{liu2015spatio} proposed a video-based pedestrian re-identification method based on the proposed spatiotemporal body-action model.
	Li~\etal~\cite{limulti} also proposed a multi-shot person re-identification method based on iterative appearance clustering and subspace learning for effective multi-shot matching.
	In addition, a multi-shot matching person re-identification using deep recurrent neural network~(RNN)~\cite{mclaughlin2016recurrent} was recently proposed. This implies that multi-shot matching with deep learning techniques will be a new trend in person re-identification.
	
	Although multi-shot matching person re-identification methods have overcome the limitations of single-shot matching to some extent, ambiguities still arise from the various viewpoints and pose changes.

		\begin{figure*}[] 
			\centering
			{\includegraphics[width=1.9\columnwidth]{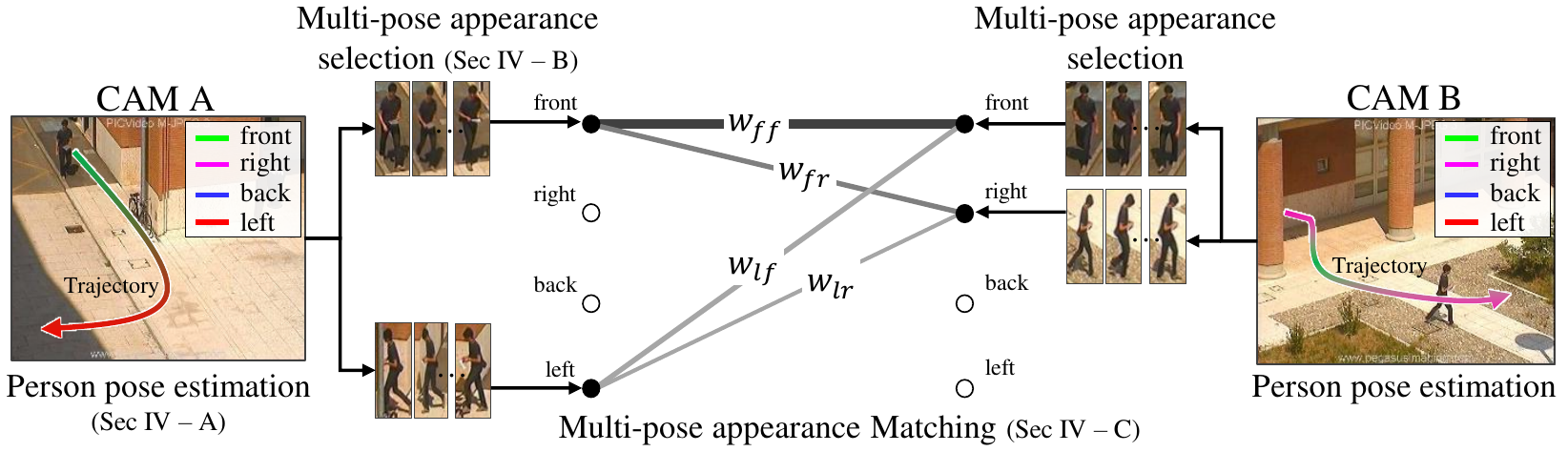} }  
			\caption{The proposed pose-aware multi-shot matching (PaMM) framework for person re-identification. First, the pose of each person is estimated. Second, a multi-pose model is generated. Finally two multi-pose models are matched based on pre-trained matching weights. The thicknesses of lines indicate the matching weights.}
			
			\label{fig:framework}
		\end{figure*}

	\section{Motivation and Main Ideas}
	\label{sec:motiv}

	As shown in Fig.~\ref{fig:chal_reid}, person re-identification is quite challenging due to camera viewpoint and person pose variations.
	However, what if the camera viewpoint and the pose priors of people in every non-overlapping camera are known in advance?
	In fact, progress in auto-calibration techniques~\cite{kusakunniran2009direct, lv2006camera} has enabled the extraction of additional cues such as camera parameters, ground plane, and the 3D position of people without any offline calibration tasks~\cite{zhang1999flexible}. Exploiting those additional cues permits the estimation of poeple's poses, as described in Section~\ref{subsec:viewpoint_est}.
	This paper fully exploits those additional cues for multi-shot matching and proposes the Pose-aware Multi-shot Matching (PaMM) for efficient person re-identification.

	Suppose that upon estimating camera viewpoints and people's poses, there is a simple 2 vs. 2 matching scenario that contains one same-pose matching (\textit{f}ront--\textit{f}ront) and three different-pose matchings (\textit{f}ront--\textit{r}ight, \textit{l}eft--\textit{f}ront, \textit{l}eft--\textit{r}ight) as shown in Fig.~\ref{fig:framework}.
	The result of the same-pose matching can generally be expected to be more reliable than those of different-pose matchings, since people maintain their appearance between cameras when their poses are the same~(This work excludes photometric issues such as illumination changes and camera color response differences).
	Then, such a multi-shot matching scenario, it is desirable that same-pose matching (\textit{f}ront--\textit{f}ront) plays a more important role than different-pose matchings. 
	Hence, this work incorporates this matching idea by aggregating the matching scores of all pose matchings in a weighted summation manner, as shown in Fig.~\ref{fig:framework}, where the thicknesses of the lines indicate the matching weights. We also study how to efficiently train matching weights and match between multi-shot appearances using pose information.

	\section{Proposed PaMM Framework}
	\label{sec:proposed}
	
	The proposed person re-identification framework, first estimates the camera viewpoint and people's poses (Section~\ref{subsec:viewpoint_est}) and then
	generate multi-pose models containing feature descriptors groups extracted from four image clusters obtained based on the person poses (\textit{i.e.} \textit{f}ront, \textit{r}ight, \textit{b}ack, \textit{l}eft) (Section~\ref{subsec:model_gen}).
	Matching scores between multi-pose models in a weighted summation manner are calculated using the pre-trained matching weights after multi-pose models are generated (Section~\ref{subsec:multi-pose matching}). 
	The matching weight training is described in Section~\ref{subsec:train_weights}.
	Fig.~\ref{fig:framework} illustrates the overall framework for the proposed person re-identification.

	\subsection{Person pose estimation}
	\label{subsec:viewpoint_est}

	Before estimating people's poses, the camera intrinsic and extrinsic parameters (or camera pose) are estimated using auto-calibration algorithms such as~\cite{kusakunniran2009direct,lv2006camera}.
	Then, the relationship between an image (pixel coordinates) and the real world (world coordinates) is described as 
	\begin{equation}
	\left[ \begin{matrix} u \\ v \\ 1 \end{matrix} \right] ={ \mathbf{ K } }\left[ \begin{matrix} { \mathbf{ R } } & { \mathbf{ t } } \end{matrix} \right] \left[ \begin{matrix} \begin{matrix} X \\ Y \end{matrix} \\ \begin{matrix} Z \\ 1 \end{matrix} \end{matrix} \right] ,
	\end{equation}
	where $\mathbf{K}$, $\mathbf{R}$, and $\mathbf{t} = {[{ X }_{ cam },{ Y }_{ cam },{ Z }_{ cam }]}^{\top}$ represent a camera intrinsic matrix, rotation matrix, and position, respectively, $[u,v]$ and $[X,Y,Z]$ represent the image and world coordinates, respectively.
	Knowing the camera parameters permits the projection of every object in an image onto the ground plane (world XY plane). 
	An object $k$ that appears in frame $t$ for camera $C$ is denoted as $\mathbf{ O }_{ t }^{ C,k }=(\mathbf{ P }_{ t }^{ C,k },\mathbf{ v }_{ t }^{ C,k }, { \theta }_{ t }^{ C,k })$, where $\mathbf{ P }_{ t }^{ C,k }=\left[ { X }_{ t }^{ C,k },Y_{ t }^{ C,k },1 \right] ,$ $\mathbf{ v }_{ t }^{ C,k },{ \theta }_{ t }^{ C,k }$ are the position, velocity, and person pose angle with respect to the camera, respectively.

	\begin{figure}[t]
		\centering
		{\includegraphics[width=0.585\columnwidth]{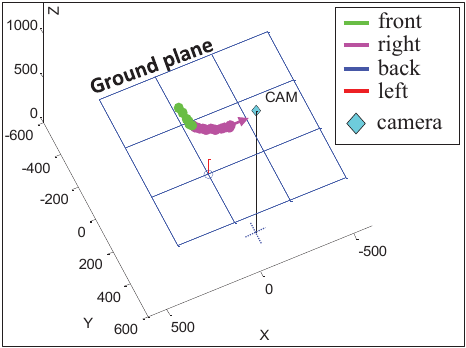}}
		{\includegraphics[width=0.395\columnwidth]{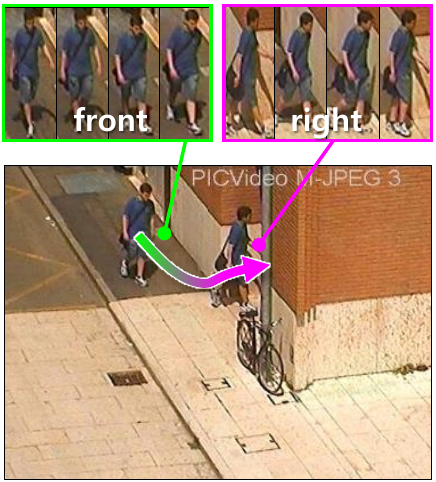}}
		\caption{Person pose estimation: (left) estimated 3D structure and person poses along the path, (right) corresponding 2D images and appearances grouped by poses.}
		\label{fig:view_estimation}
	\end{figure}
\begin{figure}[t]
		\centering
		{\includegraphics[width=0.32\columnwidth]{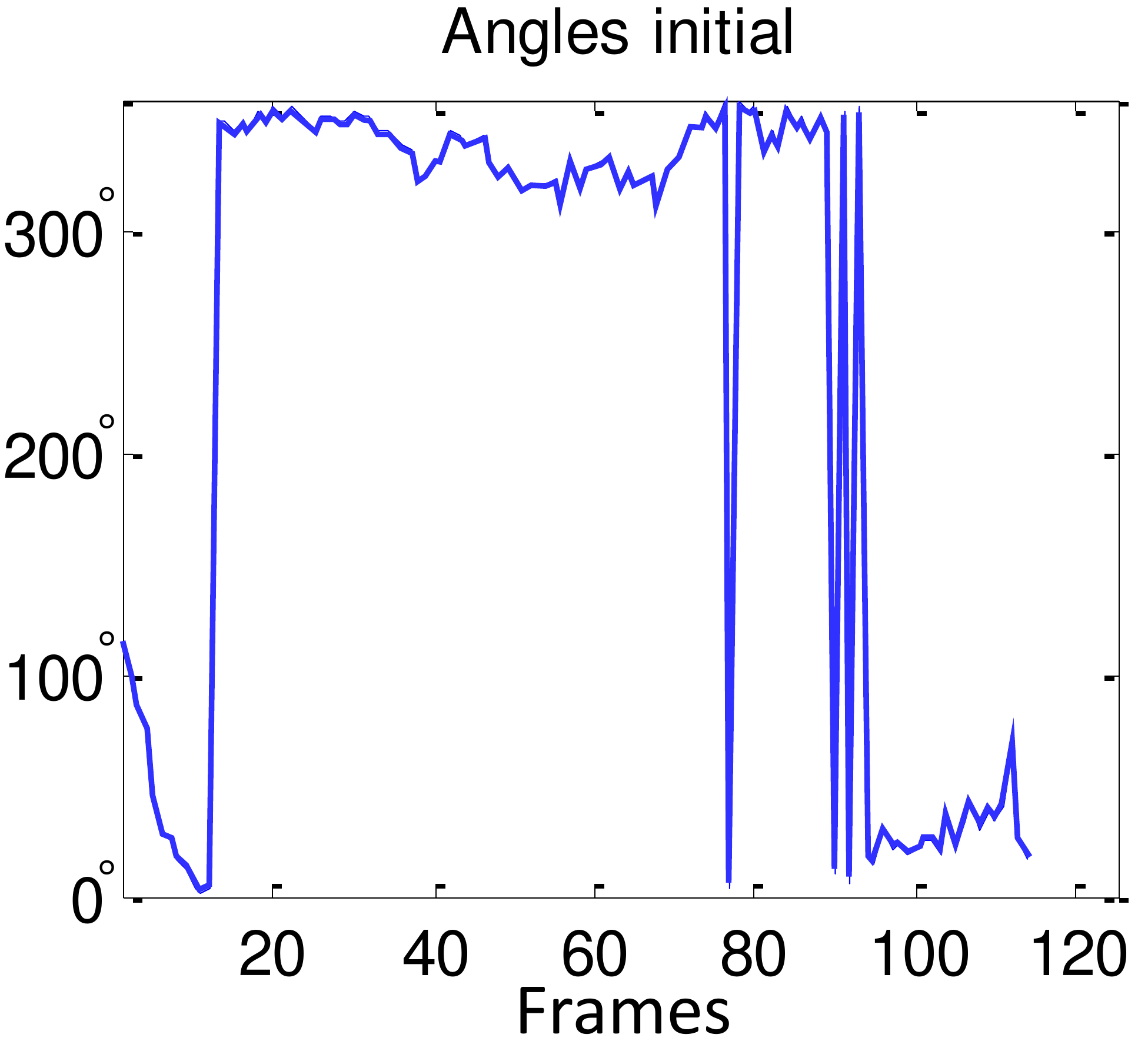}}
		{\includegraphics[width=0.32\columnwidth]{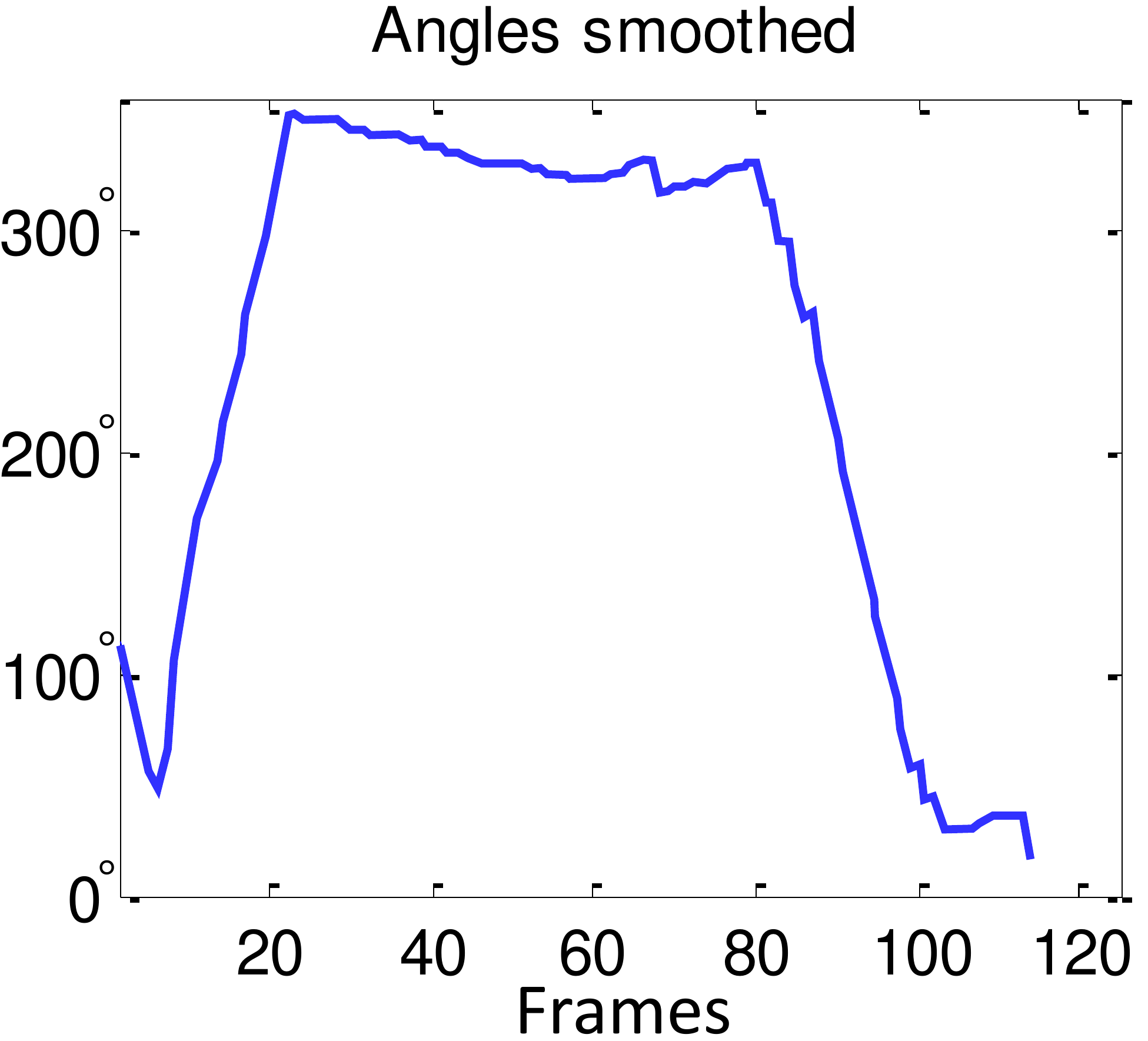}}
		{\includegraphics[width=0.32\columnwidth]{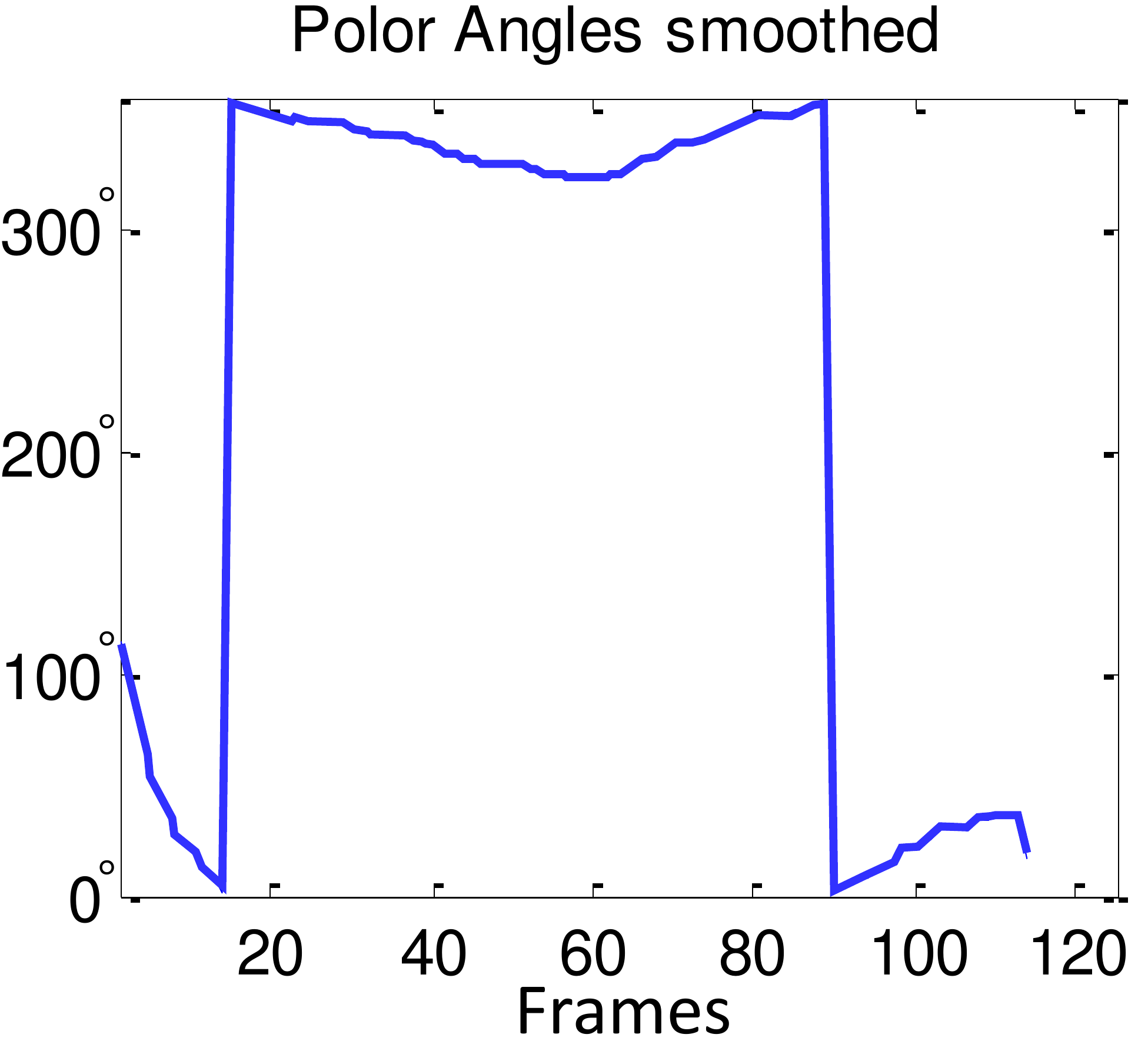}}
		\caption{(left) initial pose angle, (middle) smoothing result in Cartesian coordinates, (right) smoothing result in polar coordinates.}
		\label{fig:smoothing}
	\end{figure}

	Inspired by~\cite{wu2015viewpoint}, the velocity of a person $\mathbf{ v }_{ t }^{ C,k }$ and camera viewpoint vector $\mathbf{ c }_{ t}^{ C,k }$ are defined to estimate person's pose as 
	\vspace{0pt}\begin{equation}
	\mathbf{ v }_{ t }^{ C,k }=\left[ ({ X }_{ t+1 }^{ C,k }-{ X }_{ t }^{ C,k }),\quad({ Y }_{ t+1 }^{ C,k }-{ Y }_{ t }^{ C,k }) \right], \vspace{0pt}
	\end{equation}
	\vspace{0pt}\begin{equation}
	\mathbf{ c }_{ t}^{ C,k }=\left[ ({ X }^{ C}_{ cam }-{ X }_{ t }^{ C,k }),\quad({ Y }^{ C}_{ cam }-{ Y }_{ t }^{ C,k }) \right].\vspace{0pt}
	\end{equation}
	Assuming that pedestrians mostly walk forward, the pose angle of object $k$ can be estimated by ($C$ is omitted from here for convenience),
	\begin{equation}
	{ \theta  }^{k}_{ t }=\arccos { \left( \frac { {\mathbf{ c }^{k}_{ t }}^{\top}\cdot \mathbf{ v }^{k}_{ t } }{ \left\| \mathbf{ c }^{k}_{ t } \right\| \left\| \mathbf{ v }^{k}_{ t } \right\|  }  \right)  }.
	\label{equ:4}
	\end{equation}
	Fig.~\ref{fig:view_estimation} shows an example of estimated person poses.
	However, the initially estimated ${\theta}^{k}_{t}$ is noisy as shown in Fig.~\ref{fig:smoothing}~(a). 
	The noise is reduced by smoothing ${\theta}^{k}_{t}$ based on a moving average algorithm in the polar coordinate system as
	\begin{equation}
	{ \hat { \theta  }  }_{ t }^{ k }=\arctan { \left( { \frac { \sum _{ i=t-m }^{ t+m }{ \sin { \left( { { \theta  } }^{k}_{ i } \right)  }  }  }{ \sum _{ i=t-m }^{ t+m}{ \cos { \left( { { \theta  } }^{k}_{ i } \right)  }  }  }  } \right)  },
	\end{equation}
	where $m$ is the moving average parameter (here $m=10$). 
	Although there are several discontinuities around $0^{ \circ}$ and $360^{ \circ}$, the smoothing result is reliable due to the smoothing process in the polar coordinates, whereas the smoothing result in Cartesian coordinates is unreliable (Fig.~\ref{fig:smoothing}~(b,c)).

	\subsection{Multi-pose appearance selection}
	\label{subsec:model_gen}
	
	\subsubsection{Sample selection based on sample confidence} \label{subsubsec:sample_selection}
	Generating good multi-pose models requires filtering out unreliable samples that have incorrect angles or polluted appearances along a moving trajectory. 
	Thus, we define sample confidence to measure the reliability of person samples based on following requirements (R1--R3):

	\begin{itemize}
		\item 
		\textbf{Angle variation} (R1): We assume that the angle of a walking person will not change abruptly between temporally neighboring frames.
		If there are rapid angle changes across consecutive frames, these are regarded as unreliable samples and filtered out. We observe that, the inaccurate localization of a person generally causes a large angle variation. This is considered by measuring the angle variation as 
		\begin{equation} 
		{ \delta  }^{ k }_{ t }=\min { \left( d( { \hat { \theta  }  }_{ t }^{ k } ) ,\left| d( { \hat { \theta  }  }_{ t }^{ k } ) -360 \right|  \right) , } \vspace{0pt}
		\end{equation}
		where $d( { \hat { \theta  }  }_{ t }^{ k } ) =| { \hat { \theta  }  }_{ t-1 }^{ k }-{ \hat { \theta  }  }_{ t }^{ k } |  $. Even though there is an angle discontinuity between $0^{ \circ}$ and $360^{ \circ}$, ${ \delta  }^{ k }_{ t }$ is reliably calculated due to the second term of the $\min$ function.
		\item 
		\textbf{Magnitude of the velocity} (R2): When a person is stationary for several frames, their velocity $\mathbf{v}^{k}_{t}$ is close to 0 and the estimated person angle based on Eq.~\eqref{equ:4} becomes 
		unreliable\footnote{To estimate person angles, we assume that people mostly move forward in Section~\ref{subsec:viewpoint_est}. However, in the case of stationary person, the assumption is not satisfied. Note that the stationary people are likely to have pure rotational motion which cannot be handled by Eq.~\eqref{equ:4}.}. 
		This problem is handled by measuring the magnitude of the person's velocity as 
		${ \left\| \mathbf{ v }^{k}_{ t } \right\|  }_{ 2 }$. A sample with a small velocity magnitude is regarded as unreliable.
		\item
		\textbf{Occlusion rate} (R3): 
		When a person is occluded by others, the sample is again considered unreliable, since person's appearance is polluted.
		Occluded samples are dealt with by measuring each person's occlusion rate as
		\begin{equation}
		{ Occ }_{ t }^{ k }=\max _{ h\in \mathbf{H}^{k} }{ \left( \frac { area({ B }_{ t }^{ k }\cap { B }^{ h }_{ t }) }{ area({ B }_{ t }^{ k }) }  \right)  },  \vspace{0pt}
		\end{equation}
		where ${ B }^{k}_{ t }$ is a 2D bounding box of an object $k$ at frame $t$, ${ B }^{h}_{ t }$ is a 2D bounding box occluding ${ B }^{k}_{ t }$, and $\mathbf{H}^{k}$ is a set of object indexes occluding object $k$. As we know the 3D position of each person $\mathbf{P}^{k}_{t}$, it is easy to find $\mathbf{H}^{k}$. 
	\end{itemize}

	\begin{figure}[t]
		\subfigure[sample confidence under wrong detection]{\includegraphics[width=1\columnwidth]{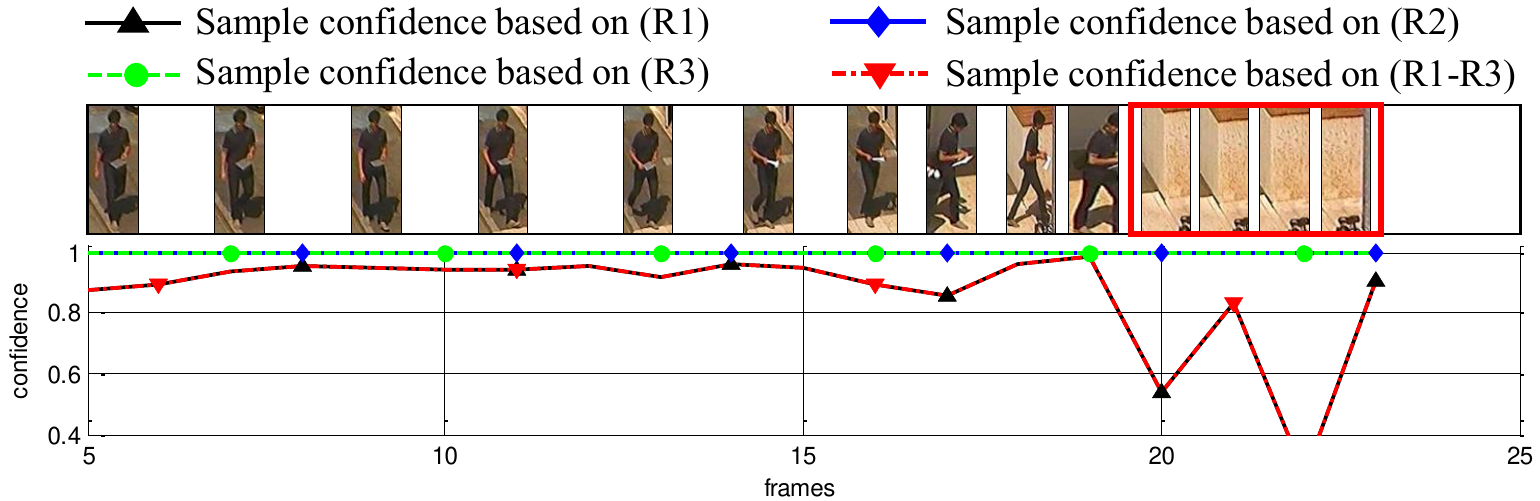} } \\ 
		\subfigure[sample confidence under pure rotation]{\includegraphics[width=1\columnwidth]{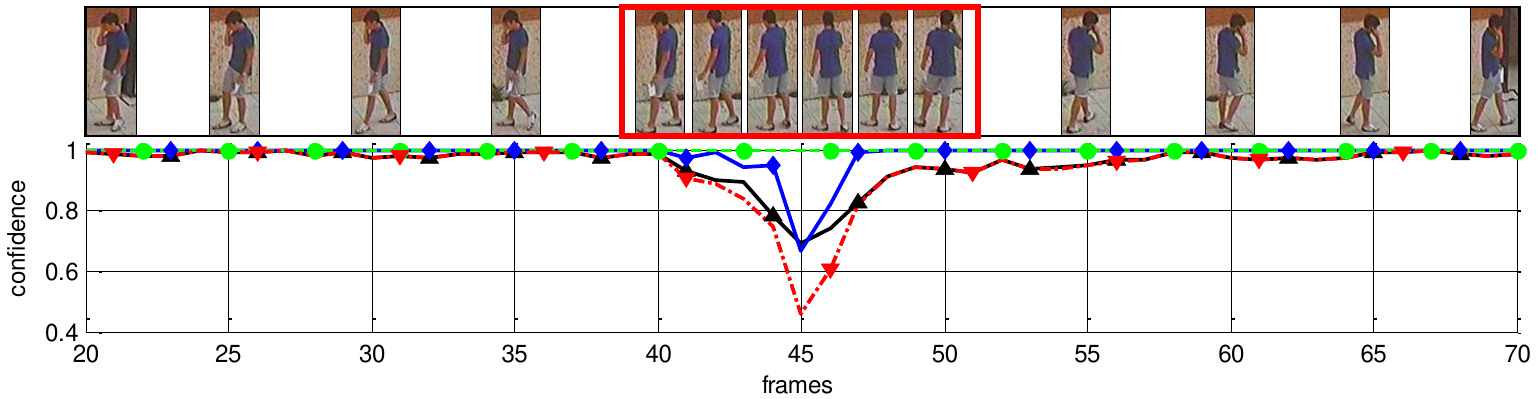}} \\
		\subfigure[sample confidence under occlusion]{\includegraphics[width=1\columnwidth]{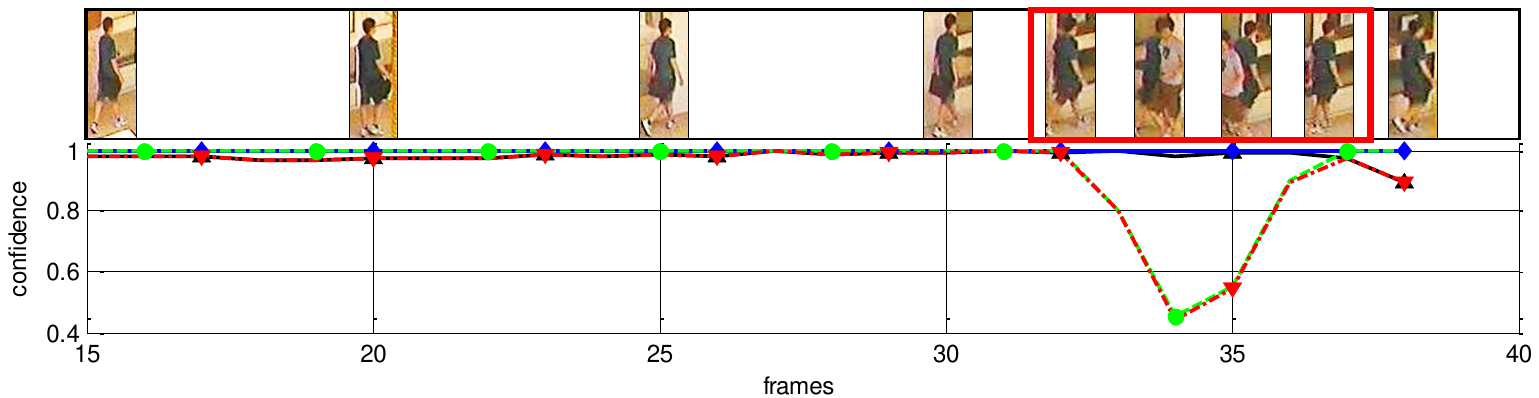}} 
		\caption{Sample confidence under various conditions (best viewed in color and at a high resolution).}
		\label{fig:sam_conf}
	\end{figure}

	\noindent Based on the above requirements, we define the sample confidence as
	\begin{equation}
	{ conf }\left( \mathbf{ O }^{k}_{ t } \right)  = { e }^{ -{ { \delta  }^{k}_{ t } } }\cdot \tanh { { \left\| \mathbf{ v }^{k}_{ t } \right\|  }_{ 2 }  } \cdot \left( 1-{ Occ }^{k}_{ t } \right).  \vspace{0pt}
	\label{eq:sample_conf}
	\end{equation}
	The sample confidence lies in [0,1]. Fig.~\ref{fig:sam_conf} shows the sample confidences under various situations. We regard a person sample as a reliable one with high sample confidence when ${ conf }\left( \mathbf{ O }^{k}_{ t } \right)>0.8$.
	
	
	\subsubsection{Generating multi-pose model}
\begin{figure}[t]
		\centering
		\subfigure[average of each cluster without sample selection]{\includegraphics[width=0.46\columnwidth]{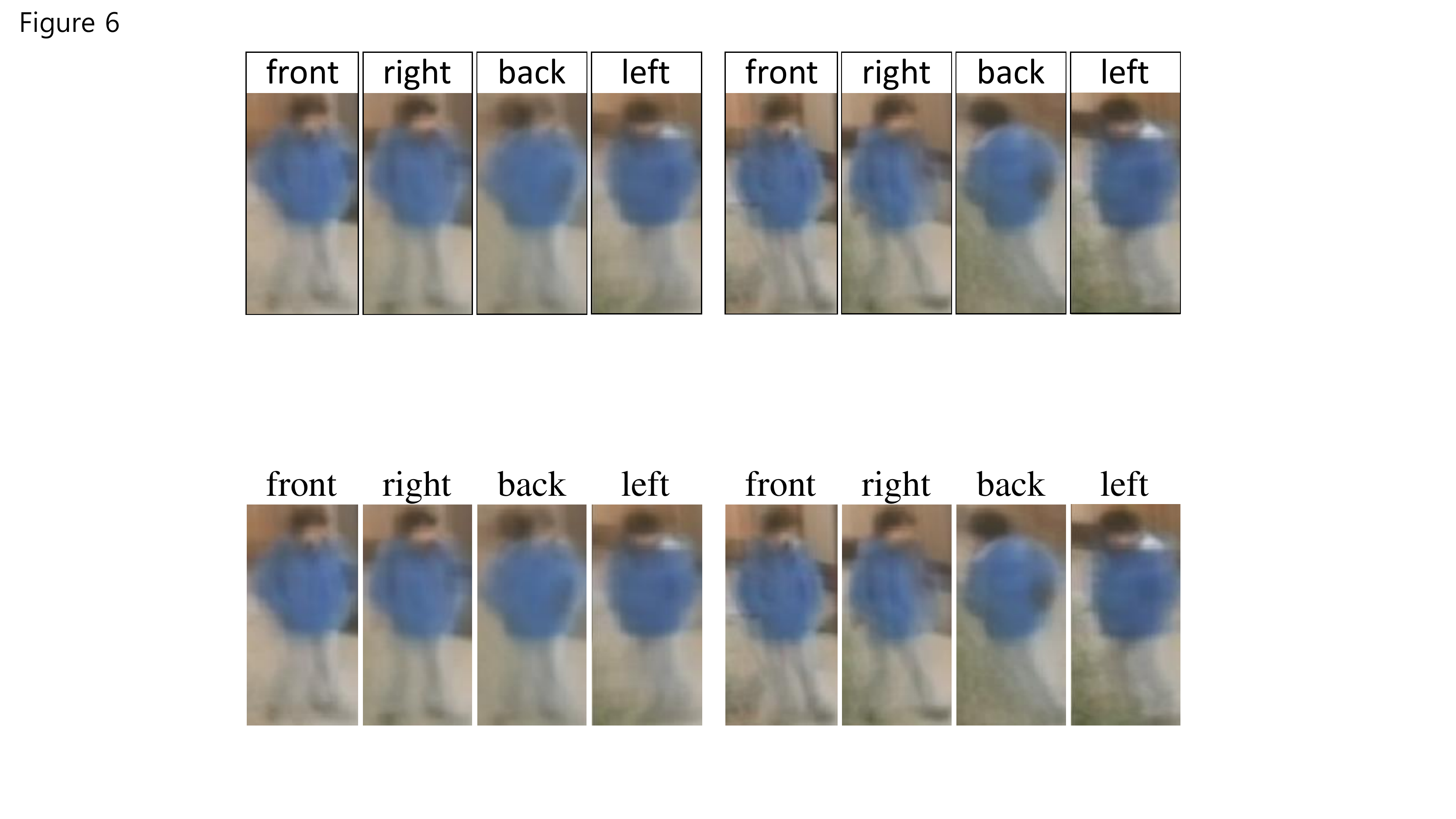}} \hspace{5pt}
		\subfigure[average of each cluster with sample selection]{\includegraphics[width=0.46\columnwidth]{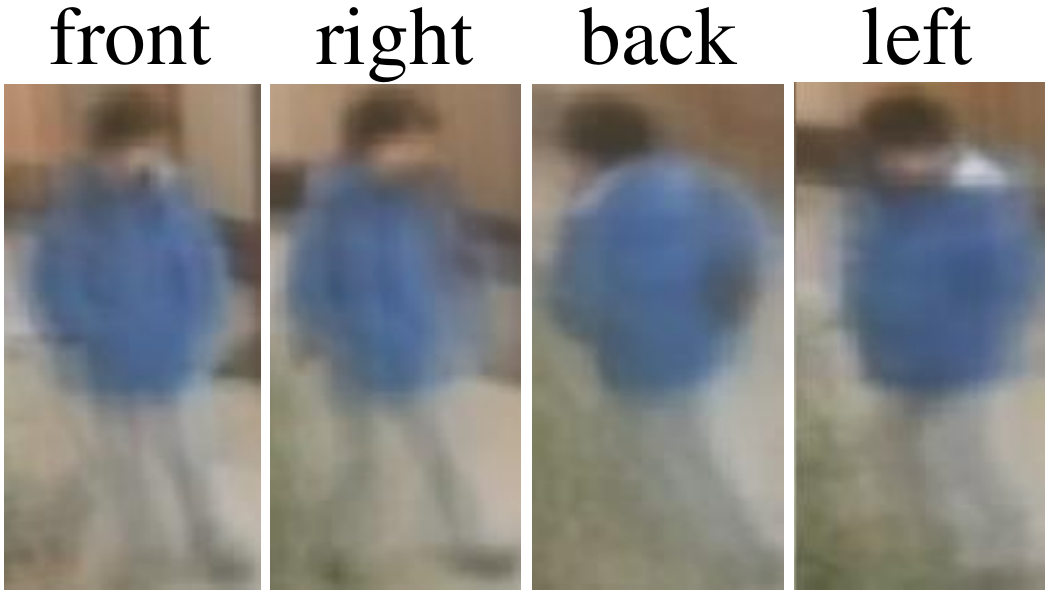}} 
		\caption{Grouping results according to person pose angles without and with sample selection. The clusters with sample selection represent more clear directivity.} 
		\label{fig:result_conf} 
	\end{figure}

	After the sample selection, we divide samples into four groups, according to their pose angles~(\textit{i.e.} \textit{f}ront, \textit{r}ight, \textit{b}ack, \textit{l}eft).
	Each group covers $90^{\circ}$ as follows:
	\begin{equation}
	\begin{split}
	& { G }_{ f }^{ k }  =\left\{ I\left(\mathbf{ O }_{ t }^{ k }\right)|{ 0 }^{ \circ  }\le { \hat { \theta  }  }_{ t }^{ k }<{ 45 }^{ \circ  },{ 315 }^{ \circ  }\le { \hat { \theta  }  }_{ t }^{ k }<{ 360 }^{ \circ  } \right\},\\
	& { G }_{ r }^{ k }  =\left\{ I\left(\mathbf{ O }_{ t }^{ k }\right)|{ 45  }^{ \circ  }\le { \hat { \theta  }  }_{ t }^{ k }<{ 135 }^{ \circ  } \right\},\\
	& { G }_{ b }^{ k }  =\left\{ I\left(\mathbf{ O }_{ t }^{ k }\right)|{ 135 }^{ \circ  }\le { \hat { \theta  }  }_{ t }^{ k }<{ 225 }^{ \circ  } \right\},\\
	& { G }_{ l }^{ k }  =\left\{ I\left(\mathbf{ O }_{ t }^{ k }\right)|{ 225 }^{ \circ  }\le { \hat { \theta  }  }_{ t }^{ k }<{ 315 }^{ \circ  } \right\},\\
	& \text{where} \qquad   t^{k}_{start} \le t \le t^{k}_{end}.
	\end{split}
	\end{equation}
	$t^{k}_{start}$, $t^{k}_{end}$ are start and end frame indexes of the object $k$, respectively. $I\left(\mathbf{ O }_{ t }^{ k }\right)$ is an image sample of the object $k$.
	It is worth noting that the proposed sample confidence~(Eq.~\eqref{eq:sample_conf}) efficiently filters unreliable samples out as shown in Fig.~\ref{fig:result_conf}.
	
	We simply represent the four groups as ${ G }_{ p }^{ k }$, where $p\in \{f,r,b,l\}$. 
	Then, an individual image that belongs to each group ${ G }_{ p }^{ k }$ is represented as
	\begin{equation*}
	{ G }_{ p_{i} }^{ k }, \quad p\in \left\{f,r,b,l \right\}, \quad 1\le i\le {N}_{p}^{k},
	\end{equation*}
	where $i$ is the index of each image in each group, and ${N}_{p}^{k}$ is the number of images in ${ G }_{ p}^{ k }$.
	For example, ${ G }_{ f_{2} }^{ k }$ denotes a second image in the group $f$ront $({ G }_{ f }^{ k })$ of the object $k$.
	
	After the sample grouping, we extract feature descriptors from the four groups. Finally, the multi-pose model of object $k$ is defined as
	\begin{equation} 
	\mathcal{ M }^{ k }={ { \left\{ f\left( { G }_{ p_{i} }^{ k } \right)| p\in\left\{ f,r,b,l \right\},  1\le i\le {N}_{p}^{k}   \right\}  } },
	\end{equation}
	where $f(\cdot)\in\mathbb{R}^{d}$ is a function that extracts a $d$-dimensional feature descriptor from an image.
 	The multi-pose model $\mathcal{ M }^{ k }$ consists of multiple feature descriptors grouped by their pose angles.
	Details of the feature extraction are described below.
	
	\textbf{Feature extraction}: Our method can apply any kind of feature descriptor extraction method. In this paper, we apply several feature extraction methods such as Histogram of Oriented Gradient~(HoG)~\cite{dalal2005histograms}, dcolorSIFT~\cite{zhao2013unsupervised} and LOMO~\cite{liao2015person} which show promising re-identification performance.
    In our feature extraction process, each person image is resized to 128$\times$48 pixels.
	Using the resized images, we extract several feature descriptors.
	
	HoG~\cite{dalal2005histograms} counts occurrences of gradient orientation on a densely sampled grid and makes an orientation histogram, and it describes the overall shape of an object well.
	dColorSIFT~\cite{zhao2013unsupervised} is a dense feature descriptor containing dense LAB-color histogram and dense SIFT. The authors pointed out that the densely sampled local features have been widely applied to matching problems due to their robustness in matching. 
	LOMO~\cite{liao2015person} analyzes the horizontal occurrence of local features and makes a stable representation by maximizing the occurrence. In addition, it handles illumination variations by applying a Retinex transform and a scale invariant texture operator.
	The results of testing the various feature extraction methods are given in Section~\ref{subsec:feature_and_metrics}.

	\begin{algorithm}[t]
		\KwData{Query objects, Gallery objects}
		\KwResult{Matching scores between queries and galleries}

		Random split training and test set;	
		
		\For{training query and gallery}{
			Extract feature descritor\;
		}
		$\mathbf{M}$ = Metric learning; {\footnotesize \% using only training set. }

		\For{test query and gallery}{
			$\theta_{t}$ = Estimate person pose\;
			$\hat{\theta}_{t}$ = Smooth person pose\;
			$conf$ = Estimate sample confidence\;
			\eIf{$conf < 0.8$}{
				Reject sample\;
			}
			{  $G_{p}$ = Perform sample grouping\;}
		}

		\For{test query and gallery}{
		$f(G_{p})$ = Extract group feature descritors\;
		$\mathcal{ M }$ = Generate multi-pose model\;
		}

		\For{test query and gallery}{
			$C\left( \mathcal{ M }^{ k }, \mathcal{ M }^{ l }\right)$ = Multi-pose model matching\;
			{\footnotesize \% $\mathcal{ M }^{ k }$ and $\mathcal{ M }^{ l }$ belong to different cameras.}
		}

		\caption{An algorithm of the propose PaMM}
		\label{alg1}
	\end{algorithm}

	\subsection{Multi-pose model matching} 
	\label{subsec:multi-pose matching}

	In this section, we describe the matching process of multi-pose models.
	Suppose that we have $\mathcal{ M }^{k }$ and $\mathcal{ M }^{ l }$, which are the multi-pose models of objects $k$ and $l$ appearing in different cameras. In order to measure the similarity between the two multi-pose models, we first calculate all pairwise feature distances between the two multi-pose models as 
	\begin{equation}
	\label{equ:distance_measure}
	{ x }_{ p_{i}q_{j} }^{k,l}= \sqrt{\left[{ f( { G }_{ p_{i} }^{ k } )  }-{ f( { G }_{ q_{j} }^{ l } )  }\right]^{\top}\mathbf{M}\left[{ f( { G }_{ p_{i} }^{ k } )  }-{ f( { G }_{ q_{j} }^{ l } )  }\right]},
	\end{equation}
	where $p,q\in\left\{ f,r,b,l \right\} $, $1\le i\le {N}_{p}^{k}$, $1\le j\le {N}_{q}^{l}$ and $\mathbf{M}$ is a $d\times d$ positive semi-definite matrix $(\mathbf{M}\preceq 0)$ learned by a metric learning algorithm\footnote{In practice, we first applied Principle Component Analysis (PCA)~\cite{jolliffe2002principal} to reduce the dimensions of the descriptors. We then performed metric learning on the PCA subspace. This is a conventional two-stage process for metric learning.}. For metric learning, we can utilize any method, such as KISSME~\cite{koestinger2012large}, ITML~\cite{davis2007information}, or LMNN~\cite{weinberger2005distance}.
	Then, the multi-pose model matching cost is computed in a weighted summation manner as
	\begin{equation}
	\begin{split}
	C\left( \mathcal{ M }^{ k }, \mathcal{ M }^{ l }\right) = \frac { \sum _{ p,q }{ { \sum _{ i,j }{{ w }_{ pq } { e }_{ pq } { x }_{ { p }_{ i }{ q }_{ j } }^{ k,l } }  } }  }{ \sum_{ p,q }{\sum _{ i,j }{ w }_{ pq } { e }_{ pq }}}, &\\
	\text{where} \qquad { e }_{ pq } = \begin{cases} 1\quad \text{if}~ (p,q)~\text{pair exists} \\ 0\quad \text{otherwise} \end{cases}, &\\
	p,q  \in \{f,r,b,l\}, \quad 1 \le i \le {N}_{p}^{k}, \quad 1\le j\le {N}_{q}^{l}. &
	\end{split}
	\label{eq:weighted sum}
	\end{equation}
	$w_{pq}$ is a matching weight that attaches importance to pairwise matching ${ x }_{ { p }_{ i }{ q }_{ j } }^{ k,l }$.
	Note that a high matching cost denotes low similarity between two multi-pose models.
	The overall algorithm of the proposed PaMM is summarized in Algorithm~\ref{alg1}.
	Matching weights training process is described in the next section.

	\subsection{Training matching weights} 
	\label{subsec:train_weights}
	
	When training the matching weights, we assume that every $p,q$ pair exists~($e_{pq}$ is eliminated). In addition, we assume that each group $G_p$ has a single image~($\sum_{i,j}$ is eliminated). Then, Eq.~\eqref{eq:weighted sum} is rewritten as
	\begin{equation}
	C\left( \mathcal{ M }^{ k }, \mathcal{ M }^{ l }\right) = \frac { \sum _{ p,q } { w }_{ pq } { x }_{ p_{1}  q_{1}  }^{k,l}   }{ \sum _{ p,q } { w }_{ pq }  }.
	\label{eq:weighted_single}
	\end{equation}
    For convenience, we also omit several indexes and terms such as object labels $(k,l)$, sample indexes $(i,j)$, and a normalization term $\sum _{ p,q }{ w }_{ pq } $ in the training step.  Then, the pairwise feature distance between two multi-pose models is simply represented as $x_{pq}$, and Eq.~\eqref{eq:weighted_single} is rewritten as
	\begin{equation}
	\begin{split}
	C\left( \mathbf{ x } \right) & =\mathbf{ w }^{ \top }\mathbf{ x },\\
	\text{where} \qquad \mathbf{ x } & = { \left\{ { x }_{ ff },{ x }_{ fr },{ x }_{ fb },\dots ,{ x }_{ ll } \right\}  }^{ \top } ,\\ 
	\mathbf{ w } & = { \left\{ { w }_{ ff },{ w }_{ fr },{ w }_{ fb },\dots ,{ w }_{ ll } \right\}  }^{ \top }. 
	\end{split}
	\end{equation}
	$\mathbf{x}\in \mathbb{R}^{16\times1}$ is a vector of pairwise feature distances and $\mathbf{w}\in \mathbb{R}^{16\times1}$ is a vector of matching weights.

	\begin{figure}[tb]
		\centering
		\subfigure[camera layout of video sets]{\includegraphics[width=0.423\columnwidth]{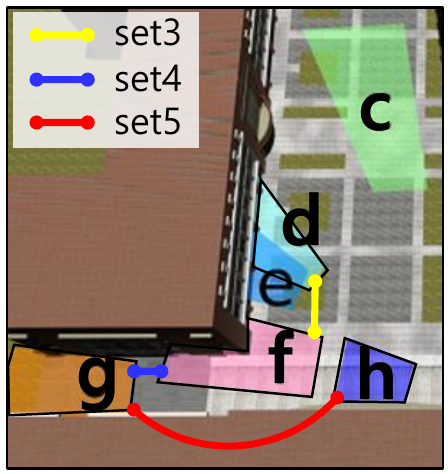}} \hspace{5pt}
		\subfigure[sample frames of each camera (d-h)]{\includegraphics[width=0.532\columnwidth]{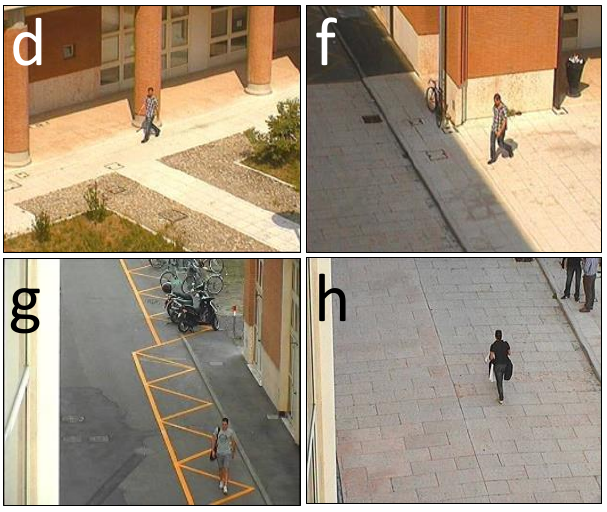}}	
		\caption{Test dataset: 3DPeS~\cite{baltieri2011_308}. We utilize three camera pair sets among the dataset (best viewed in color).}
		\label{fig:3DPeS}
	\end{figure}

\begin{figure}[t]
		\centering
		\subfigure[Examples of positive pairs]{\includegraphics[width=1\columnwidth]{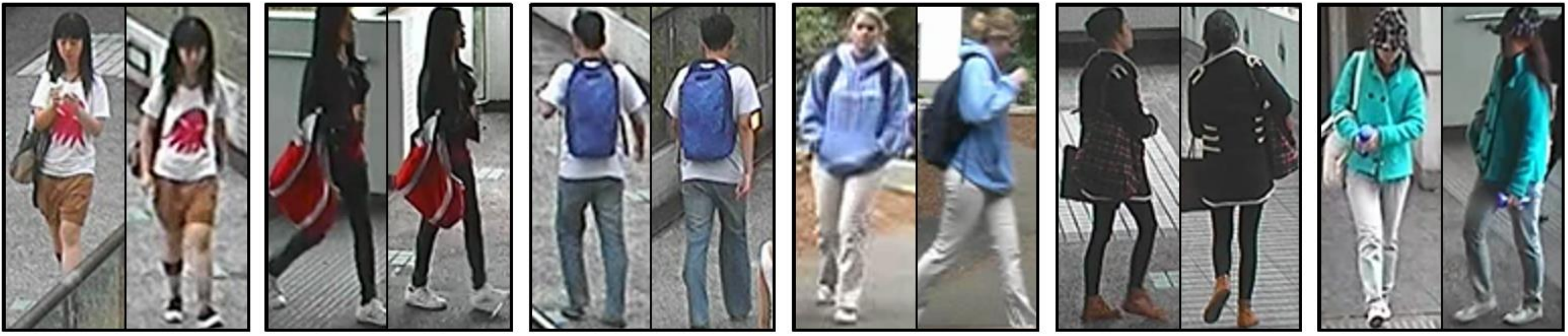}}\\
		\subfigure[Examples of negative pairs]{\includegraphics[width=1\columnwidth]{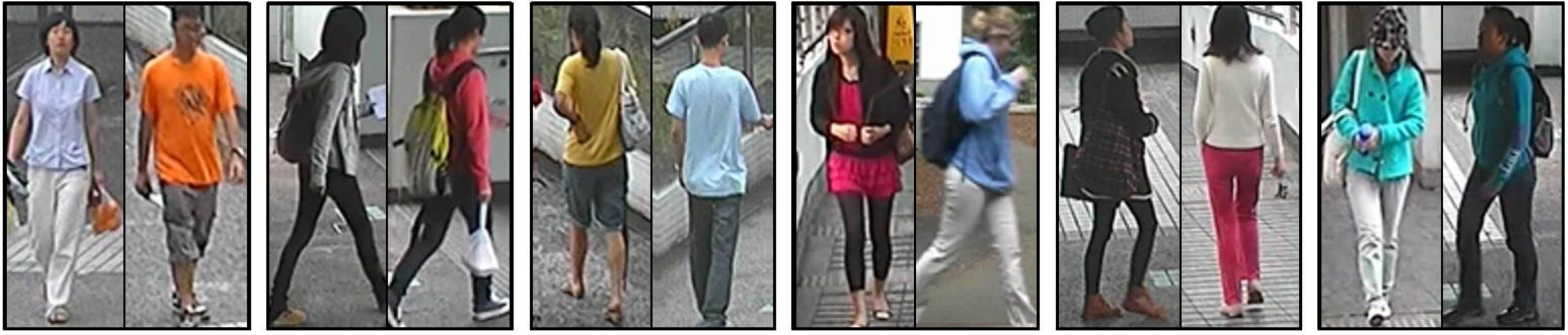}} 
		\caption{Examples of positive and negative training sample pairs.}
		\label{fig:training_sam_fig}
	\end{figure}

	In order to train matching weights $\mathbf{w}$, we collect training samples $\mathcal{D}={ \left\{ { \left( \mathbf{ x }_{ a },{ y }_{ a } \right)  }|{ y }_{ a }\in \left\{ -1,1 \right\}, 1\le a \le A  \right\}}$, where $A$ is the number of training samples and $y_{a}$ is a corresponding class of the sample.
	Fig.~\ref{fig:training_sam_fig} shows examples of the training samples (positives: $y=1$, negatives: $y=-1$).
	Given the training set $\mathcal{D}$, we exploit a Support Vector Machine (SVM)~\cite{cortes1995support} to find the weights $\mathbf{w}$ by solving the following optimization problem:
	\begin{equation}
	\begin{split}
	\underset { \mathbf{ w },\xi  }{ \text{arg min} } \left( \frac { 1 }{ 2 } { \left\| \mathbf{ w } \right\|  }^{ 2 }+\lambda \sum _{ a }^{ A }{ { \xi  }_{ a } } \right),&
	\\ s.t.~~ { y }_{ a }\left( \mathbf{ w }^{ \top }\mathbf{ x }_{ a } \right) \ge 1-{ \xi  }_{ a },~~{ \xi  }_{ a }\ge 0,
	~~\text{for}~~ 1\le& a\le A,
	\end{split}
	\end{equation}
	where $\lambda$ is a margin tradeoff parameter and ${ \xi  }_{ a }$ is a slack variable. The solution given by SVM ensures a maximal margin.
	The details and results of matching weight training are given in Section~\ref{subsec:exp_training_weights}.


	\section{Datasets and Methodology}
	\label{sec:data_metho}

	\noindent\textbf{Datasets.~}	
	For training matching weights, we used \texttt{CUHK02}~\cite{li2013locally} and \texttt{VIPeR}~\cite{gray2007evaluating}. 
	For testing methods, we used \texttt{PRID 2011}~\cite{hirzer11a}, \texttt{iLIDS-Vid}~\cite{wang2014person}, and \texttt{3DPeS}~\cite{baltieri2011_308}.

	\begin{figure*}[t]
		\centering
		\subfigure[HoG~\cite{dalal2005histograms}]{\includegraphics[width=0.63\columnwidth]{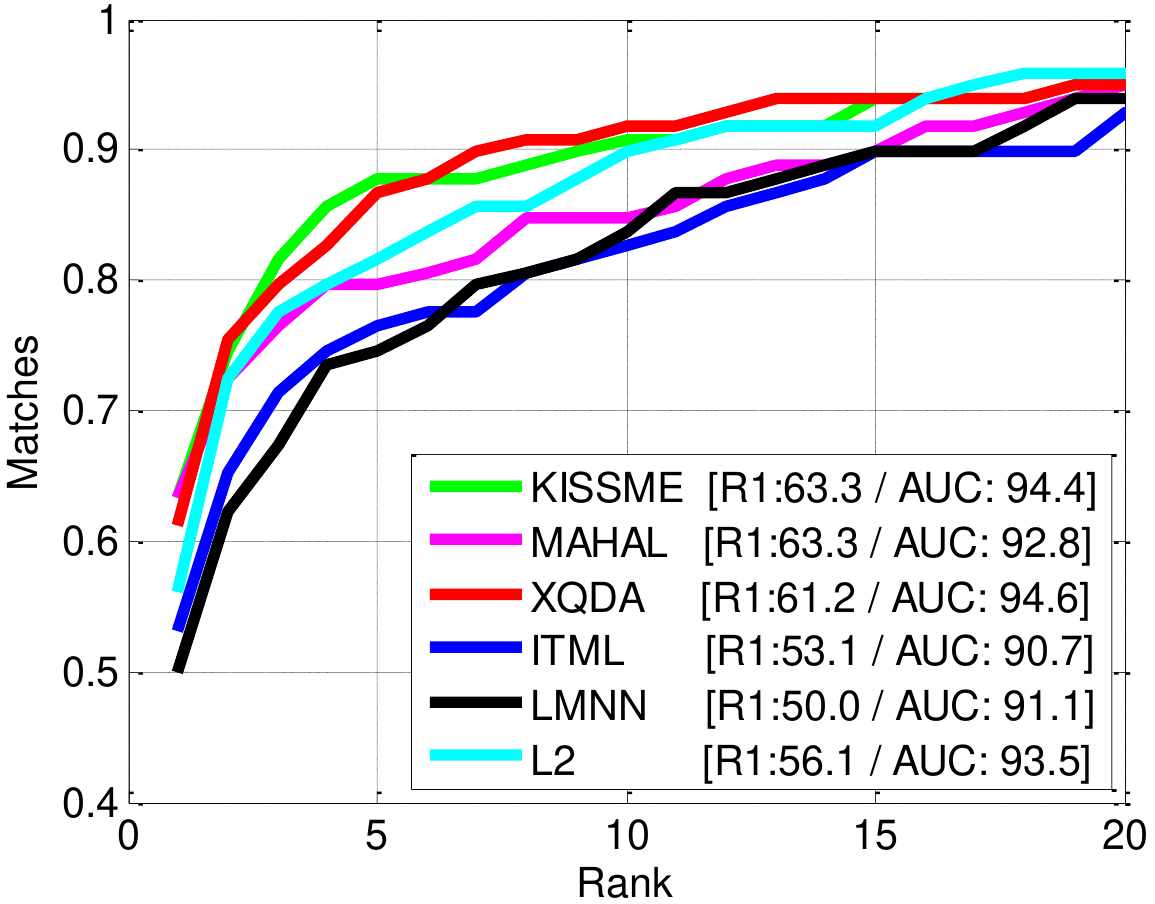}} \hspace{10pt}
		\subfigure[dcolorSIFT~\cite{zhao2013unsupervised}]{\includegraphics[width=0.63\columnwidth]{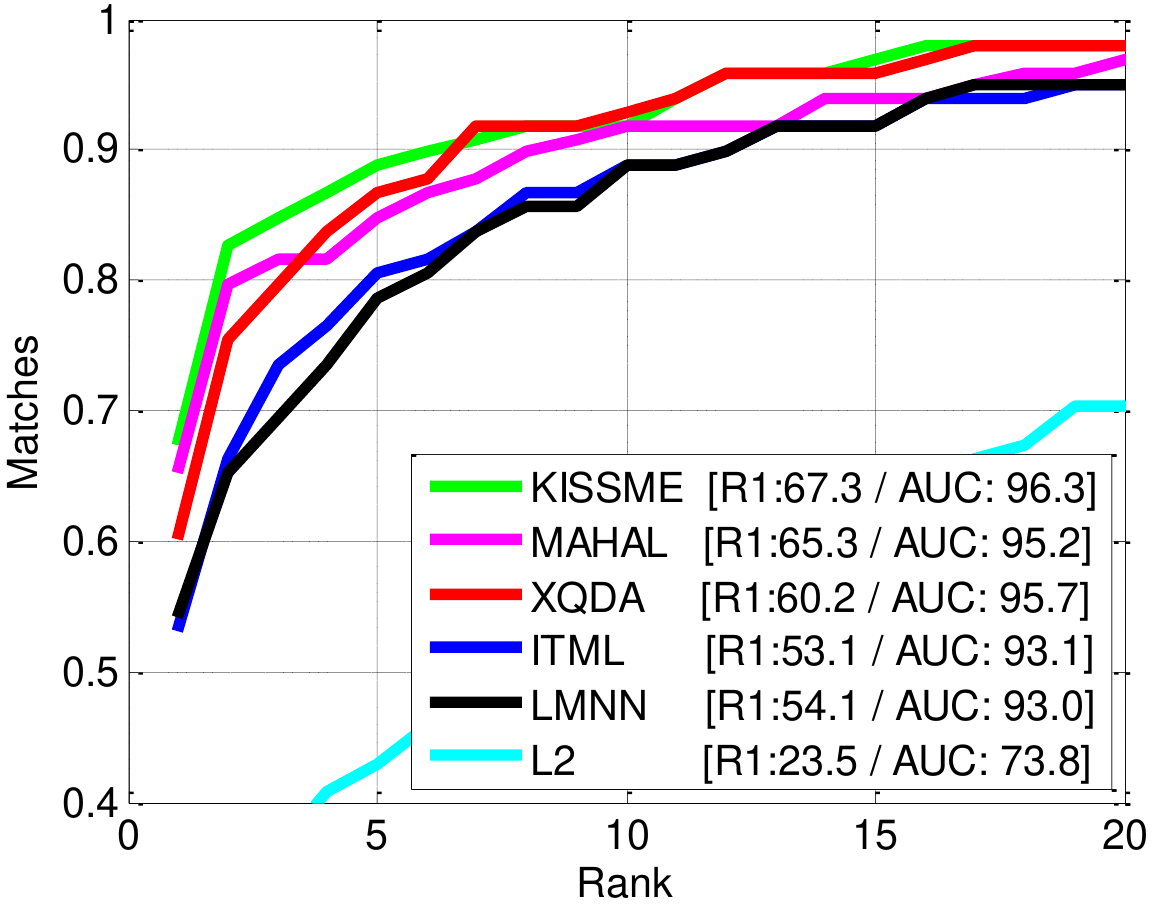}} \hspace{10pt}
		\subfigure[LOMO~\cite{liao2015person}]{\includegraphics[width=0.63\columnwidth]{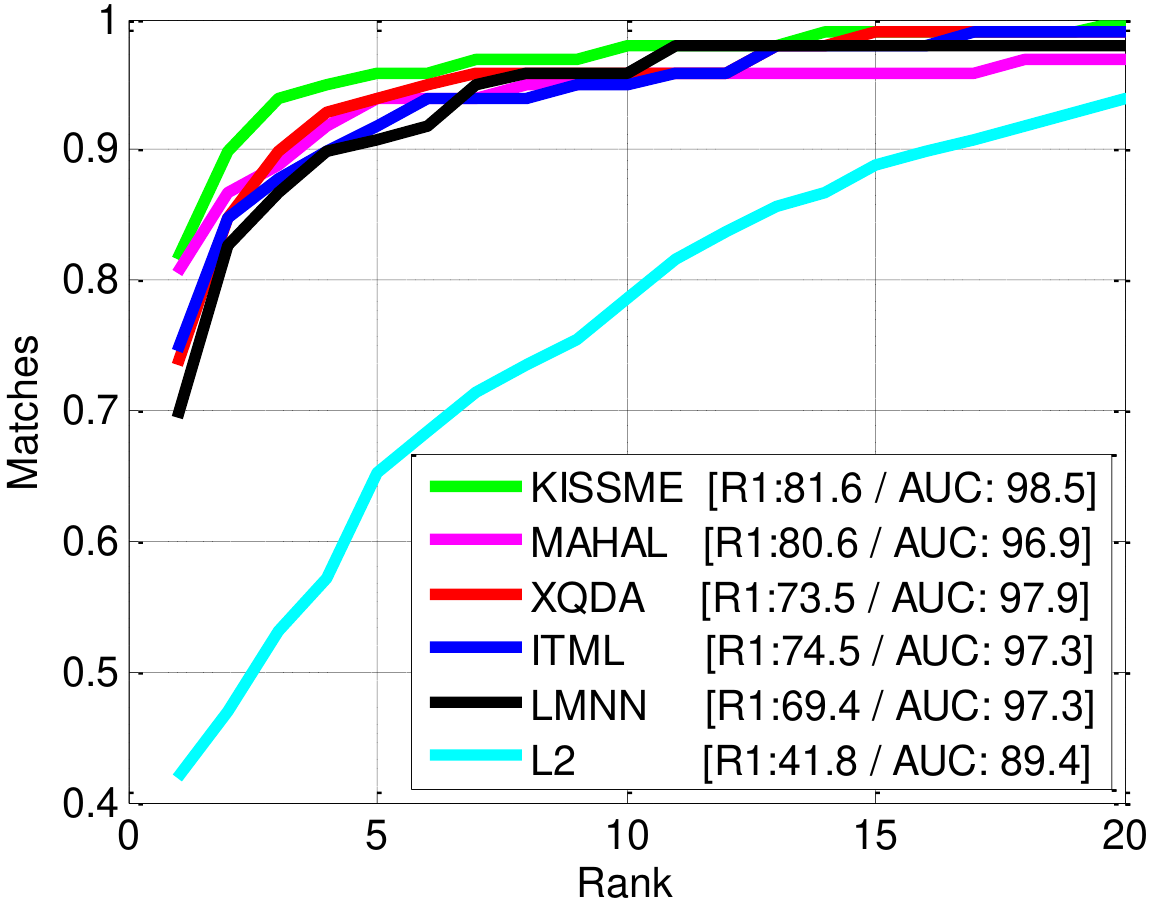}} 
		\vspace{-0pt}
		\caption{Testing of various feature extraction methods and metric learning methods. R1 denotes the rank-1 accuracy and AUC is the area under the curve of the CMC. Tested feature descriptors: (a) HoG, (b) dcolorSIFT, (c) LOMO.}
		\label{fig:test_various_feat_met}
	\end{figure*}

	\begin{figure*}[t]
		\centering
		\subfigure[$x_{ff}$]{\includegraphics[width=0.35\columnwidth]{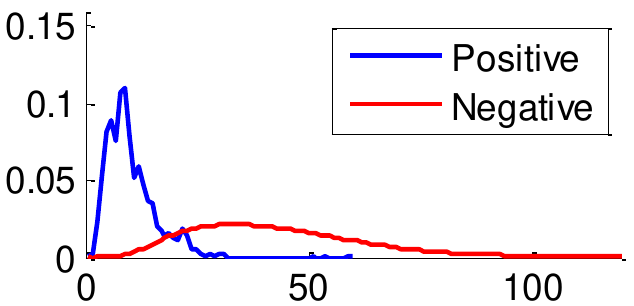}}
		\subfigure[$x_{rr}$]{\includegraphics[width=0.35\columnwidth]{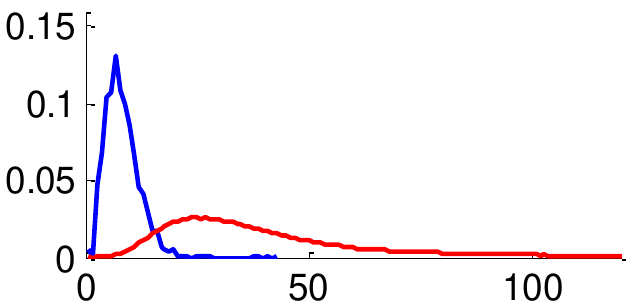}} \hspace{20pt}
		\subfigure[$x_{fr}$]{\includegraphics[width=0.35\columnwidth]{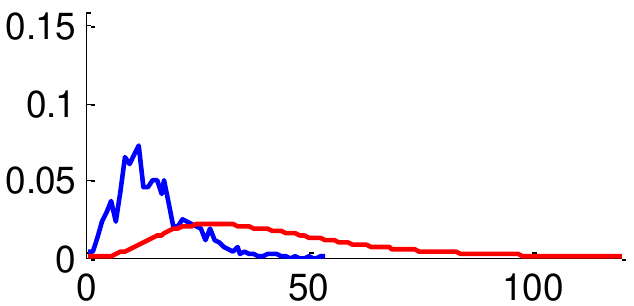}}
		\subfigure[$x_{fb}$]{\includegraphics[width=0.35\columnwidth]{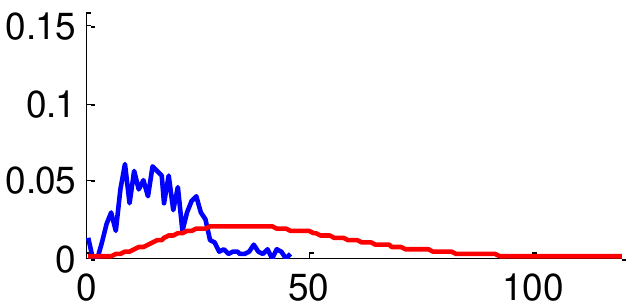}}
		\subfigure[$x_{fl}$]{\includegraphics[width=0.35\columnwidth]{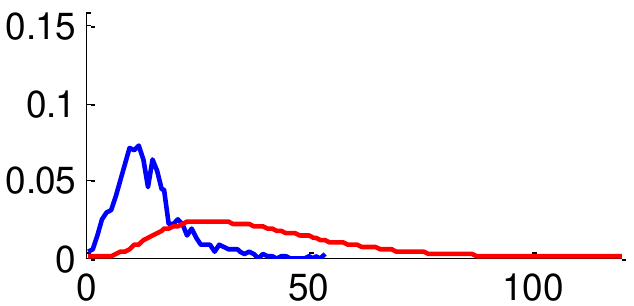}}\\
		\subfigure[$x_{bb}$]{\includegraphics[width=0.35\columnwidth]{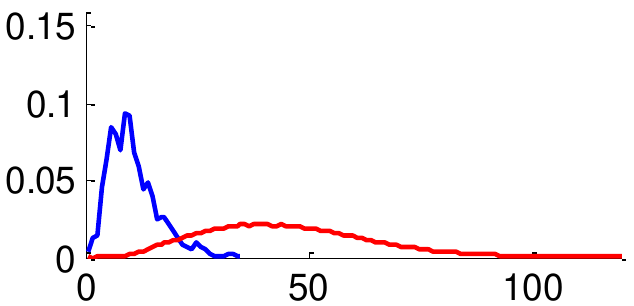}}
		\subfigure[$x_{ll}$]{\includegraphics[width=0.35\columnwidth]{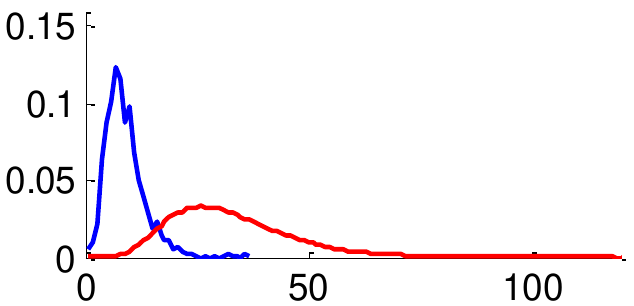}} \hspace{20pt}
		\subfigure[$x_{rb}$]{\includegraphics[width=0.35\columnwidth]{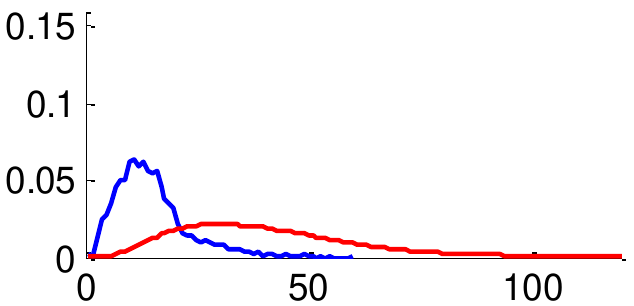}}
		\subfigure[$x_{rl}$]{\includegraphics[width=0.35\columnwidth]{./images/figure7/figure7_d.pdf}} 
		\subfigure[$x_{bl}$]{\includegraphics[width=0.35\columnwidth]{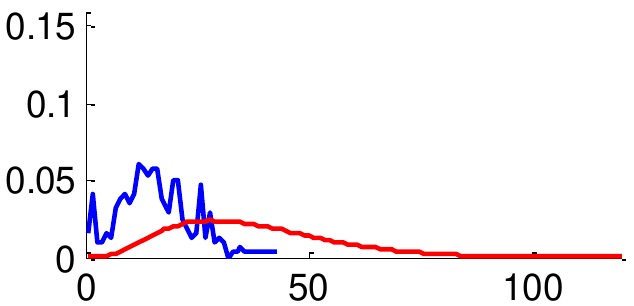}} 
		\vspace{-0pt}
		\caption{Distributions of pairwise feature distances $\left\{x_{pq}\right\}$ extracted from training data. The left two columns show the distributions of same-pose matching, while the right three columns show the distributions of different-pose matching. Tested feature descriptor: LOMO~\cite{liao2015person} and metric learning method: KISSME~\cite{koestinger2012large}.}
		\vspace{-0pt}
		\label{fig:sample_dist}
	\end{figure*}
	
	\begin{itemize}
		\item
		\texttt{CUHK02}~\cite{li2013locally} contains 1,816 people from five different outdoor camera pairs. The five camera pairs have 971, 306, 107, 193, and 239 people each with the size of 160$\times$60 pixels. 
		Each person has two images per camera that were taken at different times.
		Most people are with burdens (\textit{e.g.} backpack, handbag, or baggage).
		For our experiments, we manually extract all pose angles of each person in four directions (\textit{i.e.} \textit{f}ront, \textit{r}ight, \textit{b}ack, \textit{l}eft), since CUHK02 does not provide the pose angles. This dataset was used for training multi-shot weights $\mathbf{w}$.
		\item
		\texttt{VIPeR}~\cite{gray2007evaluating} includes 632 people and two outdoor cameras under different viewpoints and light conditions. Each person has one image per camera and each image has been scaled to be 128$\times$48 pixels. It provides the pose angle of each person as $0^{\circ}$(\textit{f}ront), $45^{\circ},$ $90^{\circ}$(\textit{r}ight), $135^{\circ}$, and $180^{\circ}$(\textit{b}ack).
		\item
		\texttt{PRID 2011}~\cite{hirzer11a} provides multiple person trajectories recorded from two different static surveillance cameras, monitoring crosswalks and sidewalks. The dataset shows a clean background, and the people in the dataset are rarely occluded. In the dataset, 200 people appear in both views. Among the 200 people, 178 people have more than 20 appearances.
		\item
		\texttt{iLIDS-Vid}~\cite{wang2014person} was created from  pedestrians in two non-overlapping cameras monitoring an airport arrival hall. It provides multiple cropped images for 300 distinct individuals and is very challenging due to clothing similarities, lighting and viewpoint variations, cluttered backgrounds, and severe occlusions.
		
		Since the datasets~(\texttt{PRID}~\cite{hirzer11a}, \texttt{iLIDS}\cite{wang2014person}) do not provide full surveillance video sequences but only cropped images, we could not automatically estimate camera viewpoints and poses of targets. Therefore, to evaluate our method with the datasets, we annotated the pose of each person manually.
		\item
		\texttt{3DPeS}~\cite{baltieri2011_308} was collected by eight non-overlapped outdoor cameras, monitoring different sections of a campus. Unlike other re-identification datasets~(\texttt{iLIDS}, \texttt{PRID}), it provides full surveillance video sequences: 
		six sets of video pairs, and uncompressed images with a resolution of 704x576 pixels at a frame rate of 15 frames per second containing hundreds of people and calibration parameters. However, this dataset provides ground-truth person identity only for selected snapshots (\textit{i.e.} no ground-truth person identity for video sequences). For our experiments, we used three sets of video pairs and manually extracted ground truth labels (identities, center points, widths, heights) of video \texttt{Set3},\texttt{4},\texttt{5}.
		We did not use \texttt{Set1},\texttt{2},\texttt{6} due to  the small number of people, and the lack of correspondences between the two cameras.
		The pose of each person was estimated as described in Section~\ref{subsec:viewpoint_est}. The camera layout and sample frames are given in Fig.~\ref{fig:3DPeS}.
		
		Even though the test datasets \texttt{Set3},\texttt{4},\texttt{5} contain people having various appearances and poses, they contain a small number of identities (\texttt{Set3}: 39, \texttt{Set4}: 24, \texttt{Set5}: 36). When the number of identities is small, the re-identification task becomes much easier because of the small pool of comparison targets. To show the person re-identification performance under more large scale data, we concatenate all datasets and generate \texttt{3DPeS-Set All} containing 99 identity pairs. It is reasonable, since the datasets~(\texttt{Set3},\texttt{4},\texttt{5}) do not share identities.
	\end{itemize}

	For readers, we open the pose annotations of \texttt{CUHK02}~\cite{li2013locally}, \texttt{iLIDS}\cite{wang2014person}, and \texttt{PRID}~\cite{hirzer11a} to the public.
	In addition, we open the ground-truth labels of~\texttt{3DPeS}~\cite{baltieri2011_308} to the public. The annotations and ground-truth labels are available online at~\url{https://cvl.gist.ac.kr/pamm/}.

	\noindent\textbf{Evaluation methodology.~}
	To compare the re-identification methods, we followed the evaluation steps described in~\cite{farenzena2010person}. First, we randomly split people identities in video pairs into two sets with equal numbers of identities, one set for training and the other set for testing. We learned several metrics, such as LMNN~\cite{weinberger2005distance}, ITML~\cite{davis2007information}, KISSME~\cite{koestinger2012large}, and Mahal~\cite{roth2014mahalanobis}, for the baseline distance functions of our person re-identification framework. After training distance metrics, we calculated all possible matches between testing video pairs. We repeated the evaluation steps 10 times. 
	We plotted the Cumulative Match Curve (CMC)~\cite{gray2007evaluating} representing the true match found within the first $n$ ranks to compare the performances of the different methods.
	Among all ranks, we mainly evaluated rank-1 accuracy which correctly finds true correspondences between two cameras.
	We also measured the Area Under Curve (AUC) of the CMC, which denotes the average accuracy of all ranks.

		\begin{figure*}[t]
			\centering
			\includegraphics[width=2\columnwidth]{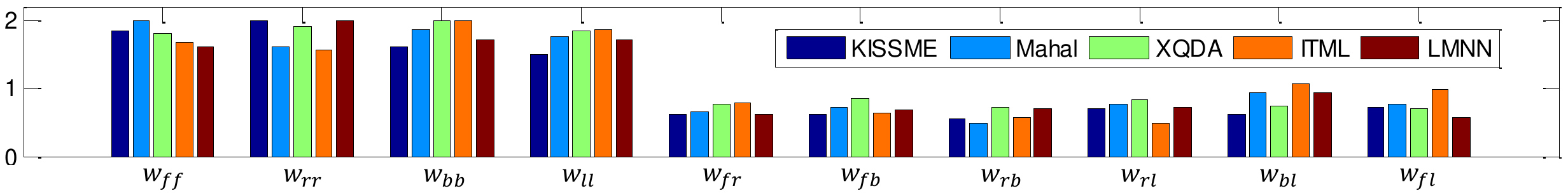}
			\vspace{-0pt}
			\caption{Results of training matching weights $\mathbf{w}\in\mathbb{R}^{10\times1}$ with LOMO~\cite{liao2015person} feature descriptor and several metric learning methods. The matching weights are equalized to lie in [0, 2].}
			\label{fig:weight_bar}
			\vspace{-0pt}
		\end{figure*}

	\section{Experimental Results}
	\label{sec:exp}
	
	\subsection{Testing various features and metric learning methods}
	\label{subsec:feature_and_metrics}
	
    We first tested various combinations of feature descriptor extraction and metric learning methods for the baseline of our person re-identification framework.
	For the feature descriptor, we tested several feature descriptor extraction methods: Histogram of Oriented Gradient~(HoG)~\cite{dalal2005histograms}, dcolorSIFT~\cite{zhao2013unsupervised}, and LOMO~\cite{liao2015person}.
	For metric learning, we tested six methods such as KISSME~\cite{koestinger2012large}, Mahalanobis~\cite{roth2014mahalanobis}, XQDA~\cite{liao2015person}, LMNN~\cite{weinberger2005distance}, ITML~\cite{davis2007information} and $L_{2}$. Note that $L_{2}$ measures the Euclidean distance between two feature vectors, \textit{i.e.}, $\mathbf{M}=\mathbf{I}$ in Eq.~\eqref{equ:distance_measure}.
	Therefore, we tested 18 combinations of feature descriptor extraction and metric learning methods.
	
	Initially, we had no trained matching weight $\mathbf{w}$ in Eq.~\eqref{eq:weighted sum}; hence, we used uniform weights as $\mathbf{w}=\mathbf{1}$ for multi-shot matching called FullMatch-avg.
	Fig.~\ref{fig:test_various_feat_met} shows the re-identification performance of each combination.
	As we can see, the combination of the feature descriptor LOMO~\cite{liao2015person} and the metric learning method KISSME~\cite{koestinger2012large} shows the best re-identification performance among the 18 combinations. 
	It shows 81.6\% rank-1 accuracy and a 98.5\% AUC score.
	In general, the LOMO feature descriptor shows promising results regardless of metric learning method (69.4\% -- 81.6\% rank-1 accuracy).
	
    We built several fusion feature descriptors, such as (HoG + dcolorSIFT), (dcolorSIFT + LOMO), (HoG + dcolorSIFT + LOMO), etc. by concatenating the feature descriptors. However, their performance was are lower than that of LOMO.
	Therefore, we utilize LOMO~\cite{liao2015person} feature as the baseline feature descriptor of our framework.

	\begin{table}[t]
		\centering
		\caption{Performance enhancement via PaMM. We used the same feature descriptor (LOMO) for all baselines. $\dag$ denotes a multi-shot matching method.} 
		\renewcommand{\arraystretch}{0.7} 
		{
			\begin{tabular}{r||c|c|c|c|c}
				\noalign{\hrule height 1pt}
				Dataset& \multicolumn{5}{c}{\textbf{\ti{3DPeS - Set All~\cite{baltieri2011_308}}}}           \\ \hline\hline
				Baseline						     		&        \multicolumn{5}{c}{KISSME~\cite{koestinger2012large}} \\  \hline
				Method $\backslash$ Rank                    & $r$ = 1  & $r$ = 3  & $r$ = 5  & $r$ = 10 & {AUC}  		   \\  \hline
				{SingleMatch}		     				    &   34.7   &  58.2    & 68.4     &  87.8    & 90.7             \\
				{MultiQ-max$^{\dag}$}			   		    &   59.2   &  75.5    & 85.7     &  93.9    & 95.4             \\
				{MultiQ-avg$^{\dag}$}			    	    &   80.6   &  91.8    &  94.9    &  98.0    & 98.5             \\
				{FullMatch-min$^{\dag}$}				    &   78.6   &\tbf{93.9}&\tbf{96.9}&  98.0    & 98.8             \\
				{FullMatch-avg$^{\dag}$}		    	    &   81.6   &\tbf{93.9}&   95.9   &  98.0    & 98.5             \\
				PaMM$^{\dag}${\scriptsize(ours)}	        &\tbf{83.7}&\tbf{93.9}&\tbf{96.9}&\tbf{100} & \tbf{99.2}       \\  \hline\hline
				Baseline						     		&         \multicolumn{5}{c}{Mahal~\cite{roth2014mahalanobis} }\\  \hline
				Method $\backslash$ Rank                    & $r$ = 1  & $r$ = 3  & $r$ = 5  & $r$ = 10 & {AUC}            \\  \hline
				SingleMatch		  				    	    &  39.8    &  61.2    & 75.5     &  86.7    &  91.8            \\
				{MultiQ-max$^{\dag}$}			    	    &   60.2   &  72.5    &   79.6   &  89.8    &  93.8            \\
				{MultiQ-avg$^{\dag}$}			     	    &   74.5   &  87.8    &   90.8   &   95.9   &  96.6            \\
				FullMatch-min$^{\dag}$				   		&   75.5   &\tbf{90.8}&   92.7   &   93.9   &\tbf{97.6}        \\
				FullMatch-avg$^{\dag}$				   		&   80.6   & 88.8     &\tbf{93.9}&\tbf{95.9}& 96.9             \\
				PaMM$^{\dag}${\scriptsize(ours)}	        &\tbf{81.6}& 89.8     &\tbf{93.9}&\tbf{95.9}&\tbf{97.6}		   \\  \hline\hline
				Baseline						     		&         \multicolumn{5}{c}{XQDA~\cite{liao2015person}  }     \\  \hline
				Method $\backslash$ Rank                    & $r$ = 1  & $r$ = 3  & $r$ = 5  & $r$ = 10 & {AUC}            \\  \hline
				SingleMatch				     				&   46.9   &  68.4    & 78.6     &  90.8    &  93.7            \\
				{MultiQ-max$^{\dag}$}			     	    &   53.1   &  71.4    &   82.7   &  91.8    &  94.2            \\
				{MultiQ-avg$^{\dag}$}			     	    &\tbf{75.5}&  87.8    &\tbf{93.9}&   96.9   &  97.9            \\	     		
				FullMatch-min$^{\dag}$						&   73.5   & 87.8     &\tbf{93.9}&\tbf{98.0}& 97.8             \\
				FullMatch-avg$^{\dag}$					 	&   73.5   & 89.8     &\tbf{93.9}&  95.9    & 97.9             \\
				PaMM$^{\dag}${\scriptsize(ours)}	        &\tbf{75.5}&\tbf{90.8}&\tbf{93.9}&\tbf{98.0}&\tbf{98.1}		   \\  \hline\hline
				Baseline						     		&         \multicolumn{5}{c}{ITML~\cite{davis2007information} }\\  \hline
				Method $\backslash$ Rank                    & $r$ = 1  & $r$ = 3  & $r$ = 5  & $r$ = 10 & {AUC}            \\  \hline
				SingleMatch				     				&   44.9   &  69.4    & 80.6     &  90.8    &  93.8            \\
				{MultiQ-max$^{\dag}$}			     	    &   58.2   &  70.4    & 78.6     &  88.8    &  93.6            \\
				{MultiQ-avg$^{\dag}$}			     	    &\tbf{75.5}&  82.7    &   89.8   &   94.9   &  96.4            \\				
				FullMatch-min$^{\dag}$						&   63.3   & 86.7     &   91.8   &   94.9   & 97.1             \\
				FullMatch-avg$^{\dag}$						&   74.5   & 87.8     &   91.8   &   94.9   & 97.2             \\
				PaMM$^{\dag}${\scriptsize(ours)}	        &   73.5   &\tbf{89.8}&\tbf{92.9}&\tbf{95.9}&\tbf{97.5}		   \\  \hline\hline
				Baseline						     		&        \multicolumn{5}{c}{LMNN~\cite{weinberger2005distance}}\\  \hline
				Method $\backslash$ Rank                    & $r$ = 1  & $r$ = 3  & $r$ = 5  & $r$ = 10 & {AUC}            \\  \hline
				SingleMatch				     				&   45.9   &  68.4    & 77.6     &  90.8    &  93.6            \\
				{MultiQ-max$^{\dag}$}			     	    &   51.0   &  72.5    &  81.6    &   89.8   &  93.7            \\
				{MultiQ-avg$^{\dag}$}			     	    &\tbf{70.4}&\tbf{87.8}&  92.9    &   96.9   &  97.4            \\				
				FullMatch-min$^{\dag}$						&   69.4   & 86.7     &\tbf{93.9}&   96.9   &\tbf{97.8}        \\
				FullMatch-avg$^{\dag}$						&   69.4   & 86.7     &   90.8   &   95.9   & 97.3             \\
				PaMM$^{\dag}${\scriptsize(ours)}	        &\tbf{70.4}&\tbf{87.8}&   91.8   &\tbf{98.0}&\tbf{97.8}		   \\  
				\noalign{\hrule height 1pt}
			\end{tabular}}
			\label{Tab1}
		\end{table}

	\subsection{Training multi-shot matching weights}
	\label{subsec:exp_training_weights}
	
	Based on Section~\ref{subsec:train_weights}, we train the matching weights $\textbf{w}$ in this section.
	In practice, we consider 10 weights rather than 16 weights due to the weight symmetry. We let ${w}_{pq}={w}_{qp}$, where $p\neq q$. Consequentially, we learn four same-pose matching weights $({w}_{ff},{w}_{rr},{w}_{bb},{w}_{ll})$ and six different-pose matching weights $({w}_{fr},{w}_{fb},{w}_{rb},{w}_{rl},{w}_{bl},{w}_{fl})$.
	
	As mentioned in Section~\ref{sec:data_metho}, in order to train the weights $\mathbf{w}\in \mathbb{R}^{10\times1}$, we use two datasets: \texttt{CUHK02}~\cite{li2013locally} and \texttt{VIPeR}~\cite{gray2007evaluating}.
	By using the datasets, we generate 3,520 positive image pairs and 35,200 negative image pairs that cover diverse pose combinations as shown in Fig.~\ref{fig:training_sam_fig}. 
	Here, a positive image pair is a pair of images of the same person and a negative image pair is a pair of images of different people regardless of the poses of people. We then extracted pairwise feature distances $\left\{ x_{ff}, x_{rr}, \dots,x_{fl}\right\} $ for all images pairs by following the metric learning steps described in Section~\ref{sec:data_metho}.
	Distributions of feature distances $\left\{x_{pq}\right\}$ are plotted in Fig.~\ref{fig:sample_dist}. For example, Fig.~\ref{fig:sample_dist} (a) shows the feature distance distribution of \textit{f}ront-\textit{f}ront image pairs of the same person (positive) and difference people (negative).
	Unfortunately, we could not make $(r,l)$ pairs using training datasets \texttt{CUHK02}, \texttt{VIPeR}, since they do not have such pairs. In order to make the distribution of $x_{rl}$, we assume that $x_{rl}$ follows a similar with similar to that of $x_{fb}$. 
	Note that a large statistical distance between positive and negative distributions implies high discriminating power.
	We observe that the same-pose matchings (Fig.~\ref{fig:sample_dist} (a,b,f,g) left two columns) are more discriminative than the different-pose matchings (Fig.~\ref{fig:sample_dist} (c-e,h-j) right three columns).
		
	After obtaining distributions of feature distances, we generate training samples $\left( \mathbf{ x }_{ a },{ y }_{ a } \right)$, where $\mathbf{ x }_{ a }\in \mathbb{R}^{10\times1}$, $y_{a} \in \left\{1,-1\right\}$ by randomly selecting each ${x}_{pq}$ from each distribution.
	Fig.~\ref{fig:weight_bar} shows the result of weight training with the LOMO~\cite{liao2015person} feature descriptor and several metric learning methods.
	The result indicates that the weights of the same-pose matchings $(ff,rr,bb,ll)$ are generally larger than those of the different-pose matchings $(fr,fb,rb,rl,bl,fl)$. The training results do not depend on the metric learning methods and show similar tendencies.
	For the following experiments, we use these trained matching weights for each baseline (feature descriptor and metric learning) individually.

		\begin{table*}[t]
			\centering
			\caption{Performance comparison result with dataset \texttt{3DPeS}. $\dag$ denotes a multi-shot matching method. The best and second best scores in each rank are marked with \tbf{bold} and \tb{blue}. AUC is an area under a curve of CMC.} 

			\begin{tabular}{l||c|c|c||c|c|c||c|c|c||c|c|c|c|c}
				\noalign{\hrule height 1pt}
				\qquad\qquad\qquad Dataset & \multicolumn{3}{c||}{\textbf{\ti{3DPeS - Set 3}}}  & \multicolumn{3}{c||}{\textbf{\ti{3DPeS - Set 4}}}& \multicolumn{3}{c||}{\textbf{\ti{3DPeS - Set 5}}} & \multicolumn{5}{c}{\textbf{\ti{3DPeS - Set All~\cite{baltieri2011_308}}}} \\ \hline
				\qquad\quad Method $\backslash$   Rank       &$r$ = 1  & $r$ = 3 & {AUC}   & $r$ = 1 & $r$ = 3& {AUC}   & $r$ = 1 & $r$ = 3 & {AUC}   & {$r$=1} & {$r$=3} & {$r$=5} & {$r$=10}& {AUC}   \\ \hline
				LOMO + $L_{2}$                               & 47.4    &  63.2   & 85.0    & 54.2    &  79.2  &  86.8   &  41.7   &  80.6   &  90.0   &  25.5   &  39.8   & 52.0    &  71.4   & 85.9    \\ 
				LOMO + KISSME~\cite{koestinger2012large}     & 36.8    &  57.9   & 80.6    & 41.7    &  66.7  &  82.6   &  41.7   &  69.4   &  87.5   &  34.7   &  58.2   & 68.4    &  87.8   &	90.7    \\ 
				LOMO + Mahal\cite{roth2014mahalanobis}       & 39.5    &  60.5   & 82.3    & 37.5    &  66.7  &  83.0   &  44.4   &  72.2   &  88.1   &  39.8   &  61.2   & 75.5    &  86.7   & 91.8    \\ 
				LOMO + XQDA\cite{liao2015person}             & 42.0    &  73.7   & 87.3    & 50.0    &  75.0  &  87.5   &  52.8   &  86.1   &  92.3   &  46.9   &  68.4   & 78.6    &  90.8   & 93.7    \\ 
				LOMO + ITML\cite{davis2007information}       & 52.6    &  76.3   & 88.4    & 58.3    &  75.0  &  85.1   &  58.3   &  80.6   &  89.0   &  44.9   &  69.4   & 80.6    &  90.8   & 93.8    \\ 
				LOMO + LMNN\cite{weinberger2005distance}     & 50.0    &  73.7   & 86.8    & 58.3    &  75.0  &  85.1   &  52.8   &  69.4   &  87.3   &  45.9   &  68.4   & 77.6    &  90.8   & 93.6    \\ \hline
				MultiQ-max$^{\dag}$					         & 55.3    &  76.3   & 90.7    & 62.5    &\tb{79.2}&  92.4   &  58.3   &  83.3   &  92.6   &  59.2   &  75.5   & 85.7    &  93.9   & 95.4    \\ 
				MultiQ-avg$^{\dag}$					         &\tb{81.6}&  94.7   &\tR{98.3}&\tb{83.3}&\tR{100}&\tb{98.3}&  69.4   &\tb{88.9}&\tR{96.3}&  80.6   &\tb{91.8}&  94.9   &\tb{98.0}& 98.5    \\ 					  
				FullMatch-min$^{\dag}$					     &   78.9  &\tb{90.0}&\tb{98.1}&\tb{83.3}&\tR{100}&  97.9   &  69.4   &\tR{90.0}&\tb{96.0}& 78.6    &\tR{93.9}&\tR{96.9}&\tb{98.0}& 98.8    \\ 
				FullMatch-avg$^{\dag}$					     &\tR{84.2}&\tR{94.7}&\tR{98.3}&\tb{83.3}&\tR{100}&  97.9   &\tb{72.2}&\tR{90.0}&\tR{96.3}&\tb{81.6}&\tR{93.9}&\tb{95.9}&\tb{98.0}& 98.5    \\
				{PaMM-ns$^{\dag}${\scriptsize(ours)}}       &\tR{84.2}&\tR{94.7}&\tR{98.3}&\tR{91.7}&\tR{100}&\tR{99.3}&\tb{72.2}&\tR{90.0}&\tb{96.0}&\tR{83.7}&\tR{93.9}&\tR{96.9}&\tR{100} &\tb{99.1}\\
				{PaMM$^{\dag}${\scriptsize(ours)}}           &\tR{84.2}&\tR{94.7}&\tR{98.3}&\tR{91.7}&\tR{100}&\tR{99.3}&\tR{75.0}&\tR{90.0}&\tR{96.3}&\tR{83.7}&\tR{93.9}&\tR{96.9}&\tR{100} &\tR{99.2}\\ \noalign{\hrule height 1pt}
			\end{tabular}
			\label{Tab2}
		\end{table*}

	\subsection{Performance enhancements via PaMM}
	\label{subsec:perform_enhance}
	
	According to the experimental result in Section~\ref{subsec:feature_and_metrics}, we utilized LOMO~\cite{liao2015person} for extracting a feature descriptor from each appearance and utilized several metric learning methods (KISSME~\cite{koestinger2012large} and others \cite{roth2014mahalanobis}\cite{liao2015person}\cite{davis2007information}\cite{weinberger2005distance}) as the baselines of this experiment.
	In this experiment, we compare the person re-identification performance based on various matching strategies (\textit{e.g.} single-shot, multi-shot, and proposed) as follows:

	\begin{figure*}[t]
		\centering
		\subfigure[\texttt{3Dpes}~\cite{baltieri2011_308}]{\includegraphics[width=0.63\columnwidth]{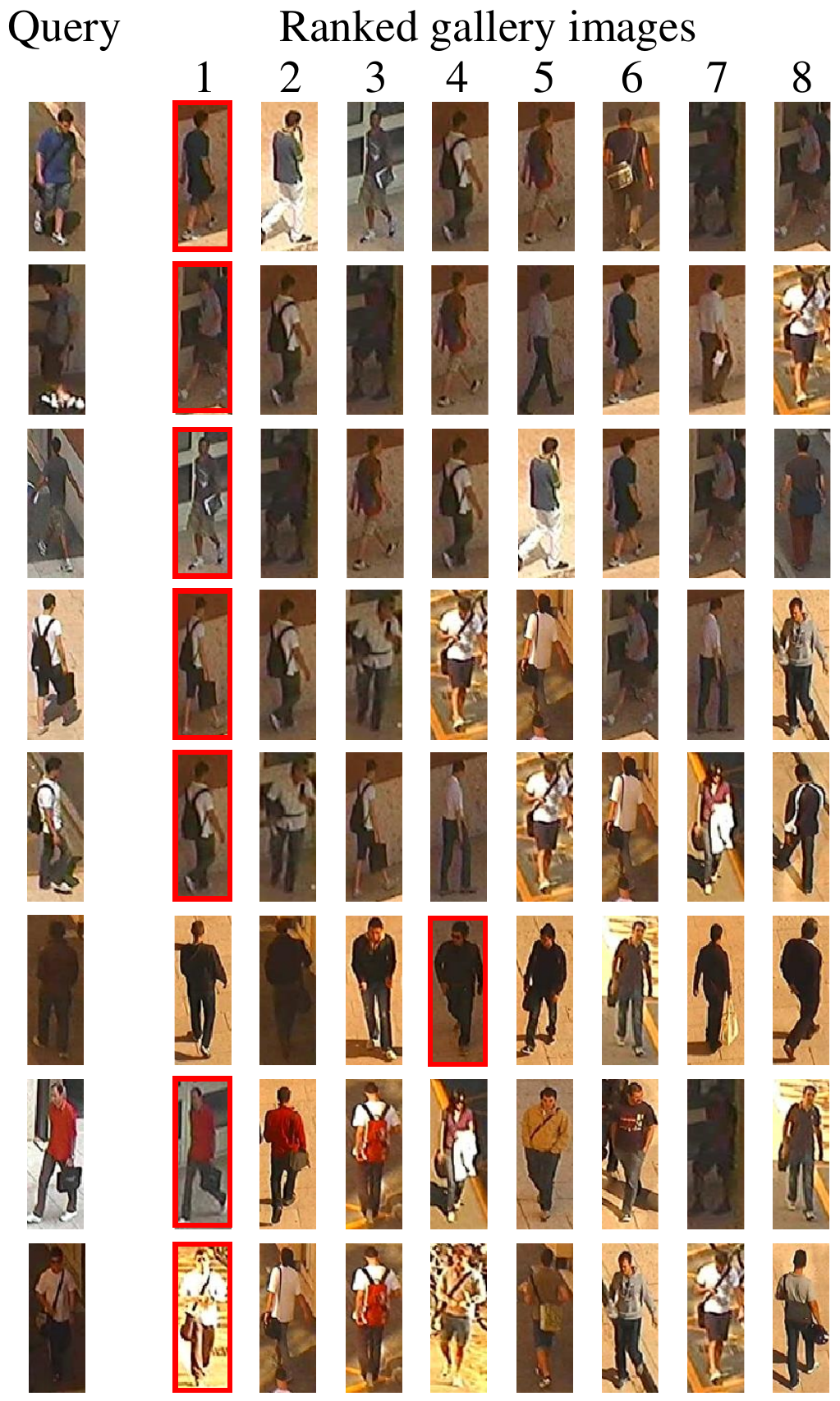}}\hspace{15pt}
		\subfigure[\texttt{PRID 2011}~\cite{hirzer11a}]{\includegraphics[width=0.63\columnwidth]{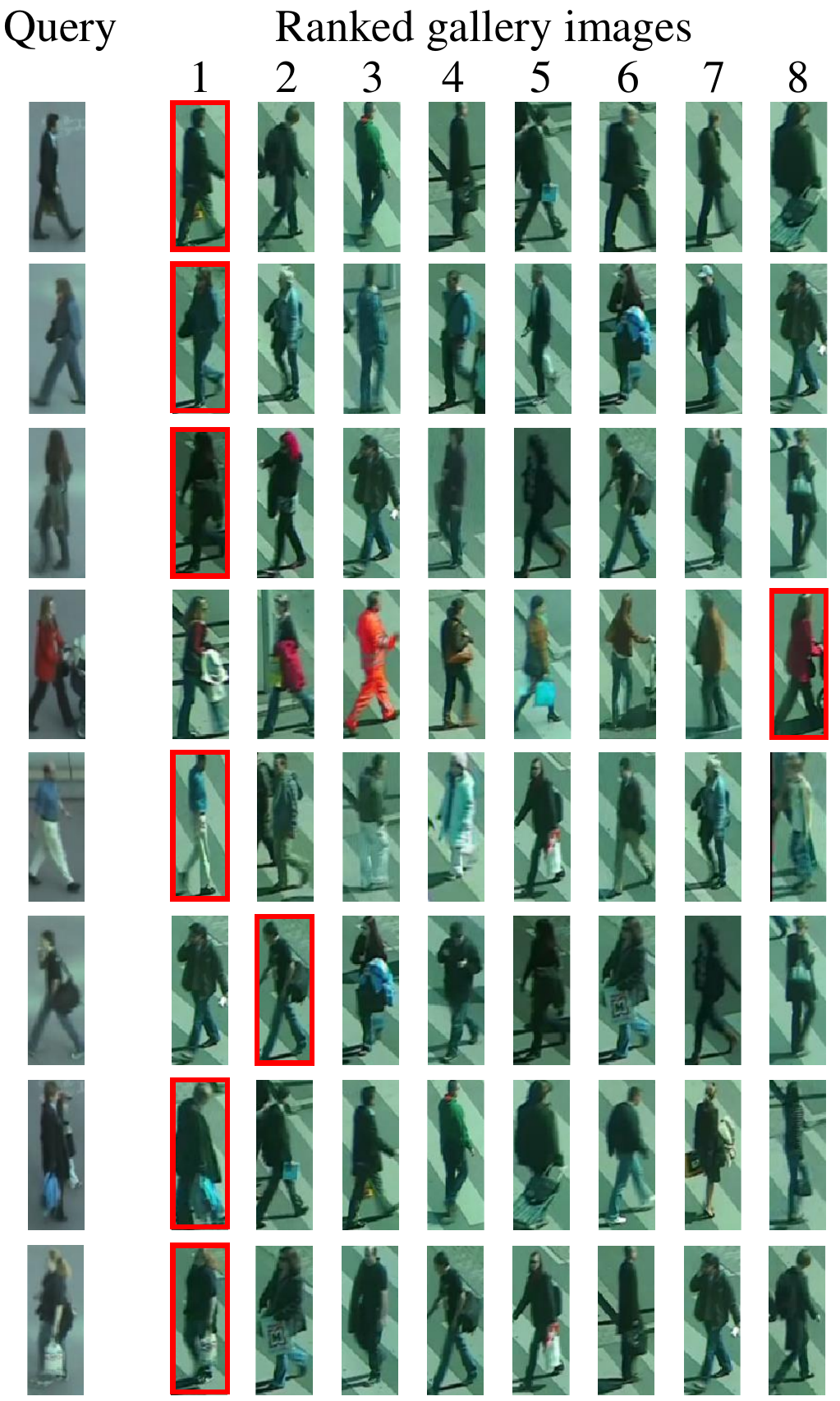}}\hspace{15pt}
		\subfigure[\texttt{iLIDS-Vid}~\cite{wang2014person}]{\includegraphics[width=0.63\columnwidth]{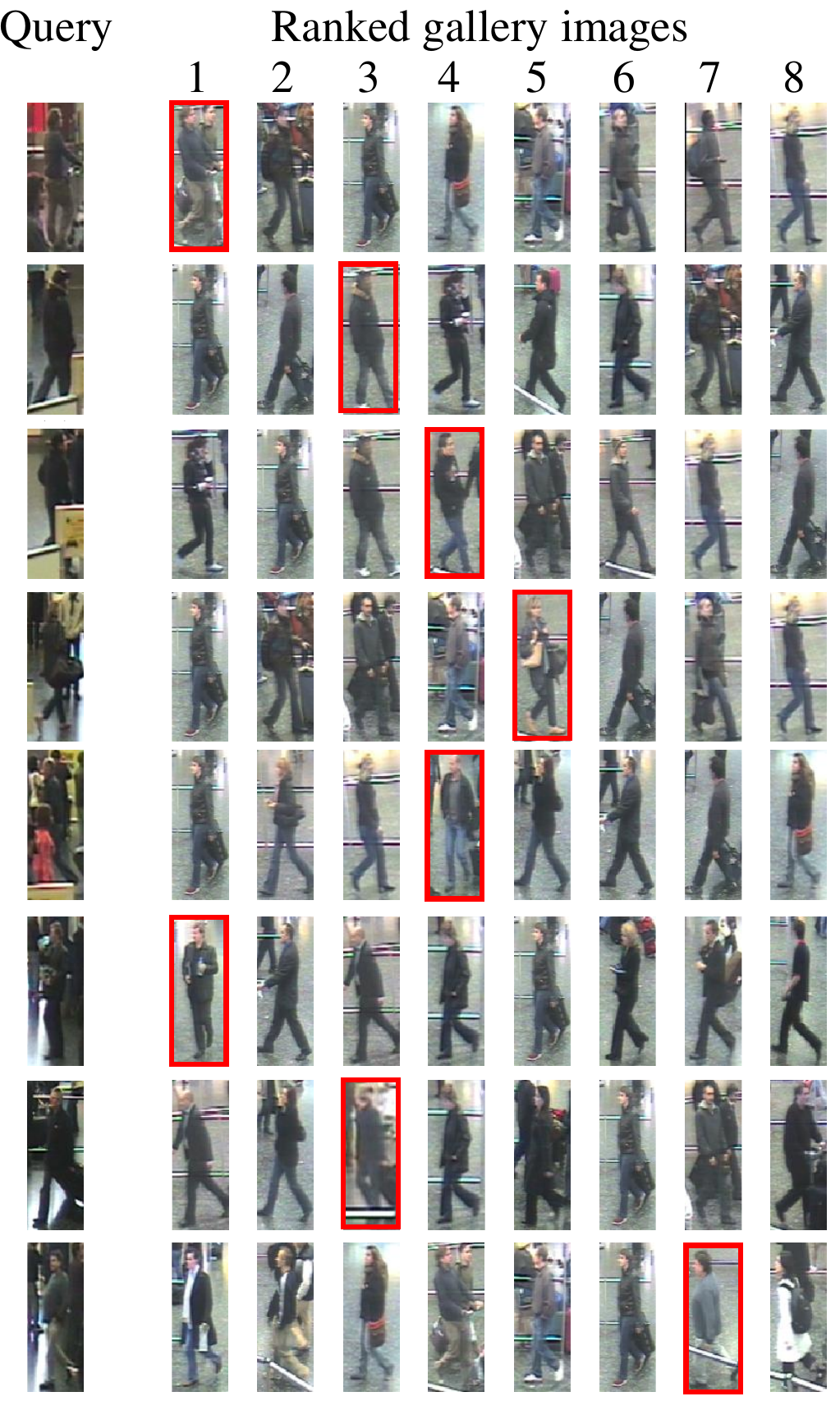}}
		\caption{Qualitative analysis results of PaMM. The true matches are highlighted with red boxes.}
		\label{fig:Ex_com_result_PaMM}
	\end{figure*}
	
	\begin{itemize}
		\item 
		SingleMatch: performing single-shot re-identification which only uses a single appearance for each person for matching. 
		As the appearance of each identity for SingleMatch, we randomly selected a single appearance for each identity. For unbiased selections, we repeated the appearance selection 10 times and calculated the average performance for the final result. 
		\item
		MultiQ-max: merging multiple appearances (\textit{i.e.} appearance feature vectors) into a single merged appearance based on the max-pooling approach: a merged feature vector takes the maximum value in each dimension from all feature vectors. 
		\item
		MultiQ-avg: merging multiple appearances into a single merged appearance based on the average-pooling approach: a merged feature vector takes the average value in each dimension from all feature vectors. 
		
		Zheng~\etal~\cite{zheng2015scalable} employed MultiQ-max and MultiQ-avg to efficiently merge multiple appearances into a single appearance and performed re-identification using the merged appearances.
		\item
		FullMatch-min: matching all possible pairs between multiple appearances and selecting the smallest matching score for the final score as used in \cite{farenzena2010person}.
		\item
		FullMatch-avg: matching all possible pairs between multiple appearances and averaging all matching scores as used in~\cite{li2015multi}.
		\item
		PaMM: proposed method.
	\end{itemize}
	
	For validating the performance enhancement via PaMM, we used the dataset \texttt{3DPeS Set All} and followed the evaluation steps explained in Section~\ref{sec:data_metho}.
	As shown in Table~\ref{Tab1}, all single-shot matching methods (SingleMatch) with different metric learning methods are improved considerably for all ranks ($r$=1,3,5,10) thanks to the proposed PaMM. The performance enhancement at $r$=1 is remarkable (achieving 24.5--49\% enhancement).
	Compared to single-shot matching methods, multi-shot matching methods (MultiQ, FullMatch, PaMM) show better re-identification performance, since they exploit several appearances for both metric learning and appearance matching.
	
	Among the various multi-shot matching strategies (MultiQ-max, MultiQ-avg, FullMatch-min, FullMatch-avg), the proposed PaMM shows the best performance regardless of the baseline: showing the best AUC scores for all baselines and showing the best rank-1 accuracies except for the ITML case. The result implies that the proposed PaMM, which exploits people's pose information, can improve the re-identification performance regardless of the baseline.
	In the consecutive experiments, we use KISSME~\cite{koestinger2012large} as the baseline metric learning method for PaMM and other multi-shot matching methods (MultiQ, FullMatch).


			
	\subsection{Test results of \texttt{3DPeS} dataset}
	\label{subsec:performance_comparison}
	
	In this experiment, we provide the detailed evaluation results of the \texttt{3DPeS} dataset. We tested \texttt{3DPeS-Set3}, \texttt{4}, \texttt{5}, \texttt{ALL} and denoted different versions of the proposed person re-identification framework as follows:
	
	\begin{itemize}
		\item
		PaMM: PaMM with all proposed methods.
		\item 
		PaMM-ns: PaMM without appearance selection.
	\end{itemize}
	As with the experiment in Section~\ref{subsec:perform_enhance}, we tested several single-shot and multi-shot matching methods. The multi-shot matching methods (Multi-Q, FullMatch, and PaMM) utilized a LOMO \cite{liao2015person} feature descriptor and a KISSME \cite{koestinger2012large} metric learning method for their baselines.
	
	Table~\ref{Tab2} shows that our methods outperform all single-shot matching and other multi-shot matching methods.
	Even though FullMatch-avg and FullMatch-min also exploit all multiple appearances of targets, the performance of both methods is lower than that of PaMM.
	This suggests that the proposed PaMM reasonably extracts key appearances among multiple appearances (Section~\ref{subsec:model_gen}) and efficiently matches multi-pose models (Section~\ref{subsec:multi-pose matching}).
	Compared to PaMM-ns, PaMM showed a slight improvement since the test dataset \texttt{3DPeS} did not encounter the challenges described in Section~\ref{subsec:model_gen}. We expect that the proposed appearance selection method in Section~\ref{subsec:model_gen} would give more performance enhancement with more challenging and complex dataset. 
	Figure~\ref{fig:Ex_com_result_PaMM} (a) shows several qualitative analysis results of the \texttt{3DPeS} dataset. As we can see, the proposed PaMM correctly finds correspondences under diverse viewpoint variations of people.
	 	

	\subsection{Test results of \texttt{PRID} and \texttt{iLIDS} datasets}
		
	We also provide evaluation results and comparisons with other state-of-the-art person re-identification methods with more public datasets such as \texttt{PRID}~\cite{hirzer11a}, and \texttt{iLIDS}~\cite{wang2014person}. In these experiments, PaMM also utilized the LOMO feature descriptor and KISSME metric learning method for its baseline.
	In the \texttt{PRID} dataset, 200 people appear in both views. We evaluated PaMM under two different test scenarios marked as $^{200}$ and $^{178}$. 
	Scenario $^{200}$ uses all 200 people pairs for testing. On the other hand, scenario $^{178}$ uses only 178 people pairs having more than 20 appearances. Most works~\cite{karanam2015person, wang2014person, wang2016person, limulti, liu2015spatio, you2016top} tested under the scenario $^{178}$, since their methods needed a sufficient video length (\textit{i.e.} multiple appearances) for extracting spatiotemporal features.

	\begin{table}[t]
		\centering
		\caption{Performance comparison result with dataset \texttt{PRID}~\cite{hirzer11a}. The best and second best scores in each rank are marked with \tbf{bold} and \tb{blue}. ${\dag}$ denotes a multi-shot matching method.} 
		{\small
			\begin{tabular}{r||c|c|c|c}
				\noalign{\hrule height 1pt}
				\qquad\qquad\qquad Dataset&  \multicolumn{4}{c}{\textbf{\texttt{{PRID 2011}}}\cite{hirzer11a}}    \\ \hline
				\qquad Method $\backslash$ Rank  & $r$=1   &$r$=5    & $r$=10  & $r$=20   \\ \hline
				$^{178}$SDALF~\cite{farenzena2010person}   	        &   4.9   & 21.5    &   30.9  &   45.2   \\ 
				$^{200}$LOMO\cite{liao2015person} + KISSME\cite{koestinger2012large} 	&   22.0  &   43.0    &   55.0  &   70.0 \\
				$^{178}$Salience~\cite{zhao2013unsupervised}	    &   25.8  & 43.6    &   52.6  &   62.0  \\
				$^{200}$LOMO + XQDA~\cite{liao2015person}		    &   39.0  & 68.0    &   83.0  &   91.0  \\    \hline
				$^{178}$SDALF$^{\dag}$~\cite{farenzena2010person}   &   5.2   & 20.7    &   32.0  &   47.9   \\ 
				$^{178}$DVR$^{\dag}$~\cite{wang2016person}          &   40.0  & 71.7    &   84.5  &   97.2   \\ 
				$^{178}$DTDL$^{\dag}$~\cite{karanam2015person}      &   40.6  & 69.7    &   77.8  &   85.6   \\ 
				$^{178}$Salience+DVR$^{\dag}$~\cite{wang2014person} &   41.7  & 64.5    &   77.5  &   88.8   \\ 				
				$^{178}$AFDA$^{\dag}$~\cite{limulti}                &   43.0  & 72.7    & 84.6    &   91.6   \\ 
				$^{200}$LOMO + XQDA$^{\dag}$~\cite{liao2015person}  &   43.0  &  82.0   &  90.0   &   98.0   \\ 
				$^{178}$TDL$^{\dag}$~\cite{you2016top}              &   56.7  & 80.0    & 87.6    &   93.6   \\ 
				$^{178}$STFV3D$^{\dag}$~\cite{liu2015spatio}        &   64.1  & 87.3    &   89.9  &   92.0   \\
				$^{200}$RNN$^{\dag}$~\cite{mclaughlin2016recurrent} &   70.0  & 90.0    &   95.0  &  97.0    \\ 
				$^{200}$PaMM$^{\dag}$ (Ours)	  		            &\tb{76.0}&\tb{94.0}&\tb{98.0}&\tb{99.0} \\ 
				$^{178}$PaMM$^{\dag}$ (Ours)	  		            &\tR{78.1}&\tR{95.5}&\tR{98.9}&\tR{100}  \\ 
				
				\noalign{\hrule height 1pt}
			\end{tabular}}
			\label{Tab3}
		\end{table}

	In Table~\ref{Tab3}, PaMM shows the best performance among ten state-of-the-art methods for all ranks while significantly enhancing its baseline performance (54\% enhancement at $r$=1).
	Although the baseline of our method (LOMO+KISSME) showed low performance, it improved significantly and showed the best performance among the state-of-the-art methods thanks to the proposed PaMM method. Compared to other state-of-the-art methods, the rank-1 accuracies of $^{200}$PaMM and $^{178}$PaMM were 6\% and 14\% higher than those of $^{200}$RNN$^{\dag}$~\cite{mclaughlin2016recurrent} and $^{178}$STFV3D$^{\dag}$~\cite{liu2015spatio}.

	Table~\ref{Tab4} shows the result of the performance comparison with dataset \texttt{iLIDS}~\cite{wang2014person}. As mentioned in Section~\ref{sec:data_metho}, \texttt{iLIDS} is much more challenging than \texttt{PRID} due to severe occlusions and lighting variations. PaMM improved its baseline performance for all ranks (46\% enhancement at $r$=1).
	Among ten state-of-the-art methods, the propose PaMM ranked second for rank-1 accuracy and third for other ranks unlike the result of \texttt{PRID} dataset.
	We think that the many severe occlusions of people broke the assumption of the proposed PaMM, because the assumption of the proposed multi-shot matching model is not satisfied under severe appearance variations.
	In particular, when the same-pose matching score is unreliable due to severe occlusions, it has a negative influence on the final multi-pose model matching score aggregation and degrades the re-identification performance.
	
	However, even though the proposed method can be influenced by severe occlusions, PaMM shows comparable performance to RNN$^{\dag}$~\cite{mclaughlin2016recurrent}, which shows the best rank-1 accuracy. The gap of rank-1 accuracy between PaMM and RNN$^{\dag}$ is $0.7$.
	In addition, when we consider both benchmark datasets \texttt{PRID} and \texttt{iLIDS}, the proposed PaMM generally shows superior performance compared to the other methods.
	It should also be noted that PaMM can achieve better performance by adopting better baseline methods.
	Several qualitative analysis results of the proposed PaMM with \texttt{PRID} and \texttt{iLIDS} datasets are illustrated in Fig.~\ref{fig:Ex_com_result_PaMM} (b,c).

	\begin{table}[t]
		\centering
	\caption{Performance comparison result with dataset \texttt{iLIDS}~\cite{wang2014person}. The best and second best scores in each rank are marked with \tbf{bold} and \tb{blue}. ${\dag}$ denotes a multi-shot matching method.} 
		{\small
			\begin{tabular}{r||c|c|c|c}
				\noalign{\hrule height 1pt}
				\qquad\qquad\qquad Dataset& \multicolumn{4}{c}{\textbf{\texttt{{iLIDS-Vid}}}\cite{wang2014person}}  \\ \hline
				\qquad Method $\backslash$ Rank              & $r$=1   &$r$=5    & $r$=10  & $r$=20  \\ \hline
				SDALF~\cite{farenzena2010person}   	         &   5.1   &  14.9   &  20.7   &  31.3   \\ 
				Salience~\cite{zhao2013unsupervised}	     &  10.2   &  24.8   &  35.5   &  52.9   \\
				LOMO~\cite{liao2015person} + KISSME~\cite{koestinger2012large} 	 &  11.3   &  27.3   &  37.3   &  49.7 \\
				LOMO + XQDA~\cite{liao2015person}		     &  18.0   &  41.2   &  54.7   &  67.0   \\ \hline
				SDALF$^{\dag}$~\cite{farenzena2010person}   &   6.3   &  18.8   &  27.1   &  37.3    \\ 
				LOMO + XQDA$^{\dag}$~\cite{liao2015person}  &   20.3  &   47.0  &  63.0   &  78.7    \\
				DTDL$^{\dag}$~\cite{karanam2015person}      &   25.9  &   48.2  &  57.3   &  68.9    \\
				Salience+DVR$^{\dag}$~\cite{wang2014person} &   30.9  &   54.4  &  65.1   &  77.1    \\ 
				AFDA$^{\dag}$~\cite{limulti}                &   37.5  &   62.7  &  73.0   &  81.8    \\ 
				DVR$^{\dag}$~\cite{wang2016person}          &   39.5  &   61.1  &  71.7   &  81.0    \\ 
				STFV3D$^{\dag}$~\cite{liu2015spatio}        &   44.3  &   71.7  &  83.7   &  91.7    \\
				TDL$^{\dag}$~\cite{you2016top}              &   56.3  &\tR{87.6}&\tR{95.6}&\tR{98.3} \\ 
				PaMM$^{\dag}$ (Ours)	  		             &\tb{57.3}&   79.3  &   87.3  &   93.3  \\ 
				RNN$^{\dag}$~\cite{mclaughlin2016recurrent} &\tR{58.0}&\tb{84.0}&\tb{91.0}&\tb{96.0} \\ 
				\noalign{\hrule height 1pt}
			\end{tabular}}
			\label{Tab4}
		\end{table}

	\section{Conclusions}
	\label{sec:conclusion} 
	In this paper, we proposed a novel framework for person re-identification, called Pose-aware Multi-shot Matching (PaMM), which robustly estimates people's poses and efficiently conducts multi-shot matching based on the pose information. We extensively evaluated and compared the performance of the proposed method using public person re-identification datasets such as \texttt{3DPeS}, \texttt{PRID 2011} and \texttt{iLIDS-Vid}. 
	
	The idea of this work is simple but very effective. 
	We showed that PaMM can improve person re-identification regardless of its baseline method. In addition, PaMM can flexibly adopt any existing person re-identification method (\textit{e.g.} feature extraction and metric learning methods) for computing pairwise feature distance in our framework. 
	The results showed that the proposed methods are promising for person re-identification under diverse person pose variances and the PaMM outperforms other state-of-the-art re-identification methods.
	We expect that PaMM will achieve much better re-identification performance when it adopts better baseline methods.

%

\ifCLASSOPTIONcaptionsoff
  \newpage
\fi



%

\bibliographystyle{IEEEtran}
\bibliography{TIP_bib}

%

\begin{IEEEbiography}{Yeong-Jun Cho} received the B.S. degree in information and communication engineering from Korea Aerospace University and the M.S. degree in information and communications from Gwangju Institute of Science and Technology (GIST), in 2012 and 2014, respectively. He is currently pursuing the ph.D. degree as a member of the Computer Vision Laboratory in GIST. His research interests include main research topics in computer vision, such as person re-identification, multi-object tracking, object detection and medical image analysis.
\end{IEEEbiography}

\begin{IEEEbiography}{Kuk-Jin Yoon}
received the B.S., M.S., and Ph.D. degrees in Electrical Engineering and Computer Science from Korea Advanced Institute of Science and Technology (KAIST) in 1998, 2000, 2006, respectively. He was a post-doctoral fellow in the PERCEPTION team in INRIAGrenoble, France, for two years from 2006 and 2008 and joined the School of Information and Communications in Gwangju Institute of Science and Technology (GIST), Korea, as an assistant professor in 2008. He is currently an associate professor and a director of the Computer Vision Laboratory in GIST.
\end{IEEEbiography}




\end{document}